\documentclass{article}
\usepackage{iclr2027_conference,times}
\iclrfinalcopy

\usepackage{amsmath,amsfonts,bm}

\def\eqref#1{equation~\ref{#1}}

\def\1{\bm{1}}

\DeclareMathAlphabet{\mathsfit}{\encodingdefault}{\sfdefault}{m}{sl}
\SetMathAlphabet{\mathsfit}{bold}{\encodingdefault}{\sfdefault}{bx}{n}

\usepackage{float}
\usepackage{hyperref}
\usepackage{url}
\usepackage{booktabs}
\usepackage{multirow}
\usepackage{graphicx}
\usepackage{amsmath}
\renewcommand{\eqref}[1]{\textup{(\ref{#1})}}
\usepackage{amssymb}
\usepackage{array}
\usepackage{tabularx}
\usepackage{centernot}
\usepackage[table]{xcolor}
\usepackage{xspace}
\usepackage{placeins}
\usepackage{wrapfig}
\usepackage{tikz}
\usetikzlibrary{arrows.meta,positioning}
\hypersetup{
  hidelinks,
  pdftitle={Beyond Corrected Memory: Execution Consistency in Multi-Agent Systems},
  pdfauthor={Zhe Yu, Zixuan Wang, Peidong Wang, Hehai Lin, Ruochen Zhao, Chengwei Qin}
}
\providecommand{\appendixheader}{}
\newcommand{\appendixpresentation}{%
  \renewcommand{\appendixheader}{\rowcolor{black!6}}%
  \arrayrulecolor{tableaccent!60!black}%
}
\newcommand{\appcontentsgroup}[2]{%
  \par\addvspace{8pt}\noindent
  \hyperref[#1]{\textcolor{tableaccent}{\textbf{\ref*{#1}\quad #2}}}%
  \hfill\hyperref[#1]{\textcolor{tableaccent}{\pageref*{#1}}}\par\nobreak
}
\newcommand{\appcontentsentry}[2]{%
  \noindent\hspace*{1em}%
  \hyperref[#1]{\ref*{#1}\quad #2}%
  \nobreak\hfill\hyperref[#1]{\pageref*{#1}}\par
}

\newcolumntype{L}[1]{>{\raggedright\arraybackslash}p{#1}}
\newcolumntype{Y}{>{\raggedright\arraybackslash}X}
\newcolumntype{R}[1]{>{\raggedleft\arraybackslash}p{#1}}
\definecolor{tableaccent}{HTML}{0F7289}
\usepackage{tcolorbox}
\tcbuselibrary{breakable,skins,listings}
\tcbset{appendixpanel/.style={
  enhanced,breakable,sharp corners,
  colback=tableaccent!2!white,colframe=tableaccent!35!black,
  borderline={0.35pt}{0pt}{tableaccent!45!black},
  colbacktitle=tableaccent!7!white,coltitle=black!85,
  boxrule=0.35pt,titlerule=0.35pt,
  fonttitle=\small\bfseries,fontupper=\small,
  boxsep=0pt,left=8pt,right=8pt,top=6pt,bottom=6pt,
  toptitle=5pt,bottomtitle=5pt,
  before skip=9pt,after skip=9pt,
  pad at break*=3pt}}
\newtcolorbox{appbox}[1]{appendixpanel,breakable=false,
  before upper=\raggedright,title={#1},
  title after break={#1\ \normalfont\itshape(continued)}}
\lstdefinestyle{appendixjson}{
  basicstyle=\ttfamily\fontsize{8.5}{10.5}\selectfont,
  columns=fullflexible,keepspaces=true,showstringspaces=false,
  breaklines=true,breakatwhitespace=false,
  numbers=left,numberstyle=\tiny\color{black!45},numbersep=8pt,
  xleftmargin=15pt,aboveskip=0pt,belowskip=0pt,
  morestring=[b]",stringstyle=\color{black!90},
  morekeywords={null,true,false},keywordstyle=\color{tableaccent}\bfseries,
  tabsize=2}
\newtcblisting{appcode}[1]{appendixpanel,breakable=false,
  colback=black!2!white,listing only,listing engine=listings,
  listing options={style=appendixjson},title={#1},
  title after break={#1\ \normalfont\itshape(continued)}}
\newcommand{\tabnote}[1]{\par\smallskip{\footnotesize\raggedright #1\par}}
\newcommand{\violation}{\textsc{Violation}\xspace}
\newcommand{\noviolation}{\textsc{NoViolation}\xspace}
\newcommand{\undetermined}{\textsc{Undetermined}\xspace}

\title{Beyond Corrected Memory: Execution\\Consistency in Multi-Agent Systems}

\author{%
\begin{tabular}[t]{@{}c@{}}
Zhe Yu\textsuperscript{1,*}\hspace{4pt}%
Zixuan Wang\textsuperscript{1,*}\hspace{4pt}%
Peidong Wang\textsuperscript{1,*}\hspace{4pt}%
Hehai Lin\textsuperscript{1}\hspace{4pt}%
Ruochen Zhao\textsuperscript{2}\hspace{4pt}%
Chengwei Qin\textsuperscript{1,\ensuremath{\dagger}}\\[5pt]
\normalfont\small\textsuperscript{1}The Hong Kong University of Science and Technology (Guangzhou)\\
\normalfont\small\textsuperscript{2}Singapore University of Technology and Design\\[3pt]
\normalfont\small\textsuperscript{*}Equal contribution.\quad
\textsuperscript{\ensuremath{\dagger}}Corresponding author.
\end{tabular}%
}

\begin{document}
\flushbottom
\setlength{\parskip}{6pt plus 2pt minus 1pt}
\makeatletter
\setlength{\@fptop}{0pt}
\setlength{\@fpsep}{12pt}
\setlength{\@fpbot}{0pt plus 1fil}
\makeatother

\maketitle
\lhead{Preprint}
\suppressfloats[t]

\begin{abstract}
Shared memory coordinates agents' actions, but correct records do not establish that those actions satisfy task requirements. Memory governance and failure diagnosis regulate or inspect recorded information; they do not by themselves establish whether it is sufficient to judge task duties. We define \emph{execution consistency} through duties governing state use, information handoffs, and final-state agreement, with explicit evidence conditions for judging fulfillment. Our core claim is that identical retained records can correspond to compliant and violating executions under the same task rule. Controlled removal of evidence such as receipt, action dependence, or response validity leaves 82.4\% of opposite-label pairs indistinguishable; restoration separates 97.9\% of the merged pairs. Natural-log annotations identify the defined violations in actual executions. However, existing logs do not always explicitly represent the execution relationships needed for these judgments. To assess the definition's practical value, we use \textbf{CAVERT}, a framework for consistency diagnosis and recovery, to extract supported relationships from logs and apply these criteria. It consistently outperforms contract-prompted LLM and rule-based baselines in diagnosis across all 12 benchmark--executor settings. Under the same gate and executor limits, it also outperforms rule-guided recovery in all four evaluated environments. These findings identify execution evidence that agent-memory and execution interfaces should preserve for reliable judgment.
\end{abstract}

\section{Introduction}
\label{sec:introduction}

Shared memory lets one agent's observations guide another's actions~\citep{yu2026multiagentmemory,margalit2026governed}. A procurement agent may complete an order after receiving notice that its approval has been revoked. The order succeeds, but the agent has not acted consistently with the updated approval. Judging this behavior requires checking how shared information was used, not just whether the task succeeded or the stored records agree. We ask \emph{what makes execution over shared memory consistent, and what evidence establishes that consistency}.

\begin{figure}[tp]
\centering
\includegraphics[width=\linewidth,pagebox=cropbox]{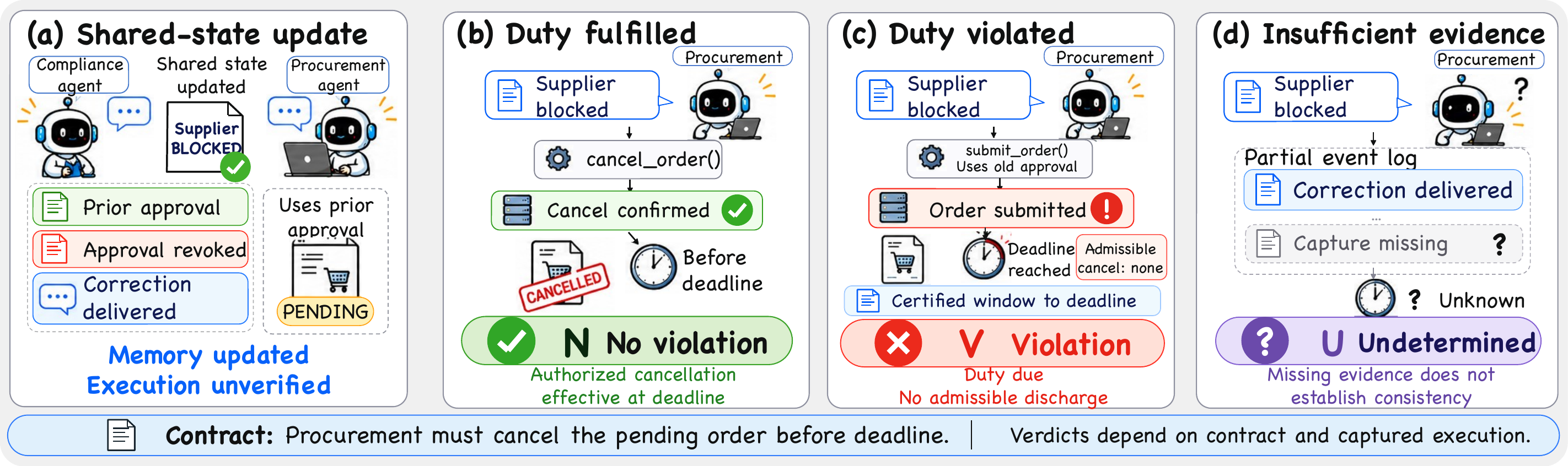}
\caption{One cancellation duty, three continuations after revocation (a): fulfillment (b, N), non-fulfillment with complete relevant records (c, V), or insufficient evidence (d, U).}
\label{fig:conflict_gap}
\end{figure}

Memory governance and transactions constrain state, dependencies, and effects~\citep{margalit2026governed,li2026memtx}. Diagnosis and process auditing evaluate available traces~\citep{barke2026agentrx,cao2026pae}, while commitment protocols and runtime verification address duties under partial observation~\citep{baldoni2015composing,mahe2022observability}. The missing link is a criterion for when an execution record contains enough evidence to distinguish duty fulfillment from violation. This requires linking task duties to message receipt, state use, tool effects, and observation coverage.

Our key insight is that \emph{record correctness does not guarantee evidence sufficiency}. A recorded correction need not establish receipt or downstream use; a cancellation request need not establish an effect lasting through the deadline (Figure~\ref{fig:conflict_gap}). Thus, \textbf{identical retained records can correspond to compliant and violating executions under the same task rule.} Better reasoning over those records cannot recover the missing distinction. Execution consistency therefore treats evidence sufficiency---not record agreement alone---as part of the judgment problem.

We define \emph{execution consistency} as judging task-duty fulfillment from available execution evidence. A judgment must establish the applicable duty, sufficient observation of its execution window, and whether an admissible response fulfilled it; unresolved evidence yields an undetermined outcome rather than presumed compliance or violation. We instantiate this principle across state use, information handoffs, and final-state agreement, covering twelve mechanisms.

To test whether accurate shared-memory contents suffice for judgment, we remove and restore selected execution evidence across 450 controlled pairs, holding each pair's contract, query, and full-execution labels fixed. Removal can merge compliant and violating executions into identical retained inputs; restoration separates nearly all merged pairs. Independent annotations of 360 natural executions confirm that the defined violations occur in practice.

However, these relationships are not always explicit in existing logs. \textbf{CAVERT}, a framework for consistency diagnosis and pre-commit intervention, extracts supported relationships, checks source references, and applies the criteria through deterministic rules, retaining unresolved evidence as unknown. Its findings guide recovery. It consistently outperforms contract-prompted LLM and rule-based diagnostic baselines across all 12 benchmark--executor settings, and rule-guided recovery across all four evaluated environments under the same gate and executor limits.

Our contributions are:
\begin{enumerate}
\item We \textbf{define task-level execution consistency} through duties and evidence of fulfillment, with explicit rules for violation, no violation, and undetermined judgments.
\item We \textbf{establish when retained records cannot distinguish compliant from violating runs}, test the required relationships through controlled removal and restoration, and identify the defined violations in natural logs.
\item We \textbf{apply these criteria through CAVERT} to diagnose existing logs and guide pre-commit intervention, demonstrating their practical value through diagnosis and recovery experiments.
\end{enumerate}

\section{Related Work}
\label{sec:related}

\paragraph{State governance.}
Which state should agents retain and be allowed to use? Distributed consistency constrains operation order, update visibility, and replica convergence \citep{herlihy1990linearizability,lloyd2011cops,vogels2009eventually,shapiro2011crdt}. Agent-memory research studies retention, access, conflicting records, and visibility \citep{hu2025memoryagentbench,margalit2026governed,ren2026gatemem,volkov2026statefuse,yu2026multiagentmemory}. TOKI formalizes write-time conflict resolution \citep{wang2026toki}; \citet{khan2026concurrency} verifies a concurrency hierarchy under deterministic generation. MemTX adds belief commit, action gating, and cascading repair \citep{li2026memtx}. Two runs can share the same approval-register history yet differ in whether the order is canceled. We define the task duties and execution evidence that distinguish such runs, linking shared state to actions, handoffs, and outcomes (Appendix~\ref{app:distributed_comparison}). RQ3 compares CAVERT with MemTX.

\paragraph{Diagnosis and process auditing.}
Which behavior explains a failure or violates a task requirement? Failure analysis classifies errors and locates responsible agents and steps \citep{cemri2025mast,zhang2025whowhen,zhang2026agentracer}. AgentRx checks constraints for attribution \citep{barke2026agentrx}; DoVer tests failure hypotheses through interventions \citep{ma2026dover}. PAE audits procedural integrity, including agreement between action claims and tool execution \citep{cao2026pae}. We focus on shared-memory duties and test whether the retained evidence distinguishes their fulfillment from violation, complementing diagnosis and process audit.

\paragraph{Obligation and behavior verification.}
Does observed behavior satisfy an explicit rule? Commitment protocols and runtime verification formalize duties, deadlines, and three-valued judgments \citep{singh1999commitments,mallya2006commitment,fornara2011obligations,bauer2011runtime}. They also analyze how partial observation affects compliance checking \citep{baldoni2015composing,mahe2022observability}. Application-level checks enforce or repair constraints \citep{balegas2015indigo,chang2025sagallm}; other work checks reasoning, plans, policy compliance, and task actions \citep{feng2026vericot,wu2026virf,levy2026stwebagentbench,froger2026gaia2}. Building on these semantics, we specify how generated messages, state use, and tool effects establish task-duty fulfillment. Controlled pairs test which relationships must be retained for judgment (Section~\ref{sec:contracts}; Appendix~\ref{app:agent_verification_comparison}).

These lines of work answer different questions. Memory consistency governs what state is retained, exposed, or eligible for use; diagnosis and runtime verification reason about failures or duties from available traces. Execution consistency asks whether the retained trace contains the relationships needed to distinguish fulfillment from violation under a task duty. Our removal--restoration study tests this distinction directly: removing those relationships can make oppositely labeled executions observationally identical to the judge.

\section{Execution Contracts and Evidence}
\label{sec:problem}
\label{sec:contracts}

Execution consistency is judged relative to a task duty and the evidence available to assess it. It does not require global agreement among agents or memory states, only enough evidence to determine whether the relevant duty was fulfilled, violated, or remains unresolved. The unit of judgment is a task duty within an execution; trace-level consistency aggregates the resulting findings.

The \emph{public contract} $P$ encodes task instructions and declared task or framework policies, fixed before evaluation. For procurement, receipt of an approval revocation activates the duty to cancel the affected order by the final checkpoint. Thus $P$ specifies responsibility, activation, scope, permitted response, and deadline; execution evidence establishes fulfillment.

\paragraph{Three contract families.}
We group duties by where consistency must hold: at state use, across information handoffs, or among designated final states.
\emph{Strict-temporal} contracts govern state use and contradiction resolution at an action or checkpoint. Conflicting writes commit ``approved'' despite a visible cancellation. Stale dependency use places an order using a replaced approval. Unresolved contradictions leave both states active at a required checkpoint. Contract-permitted exceptions, such as version pinning, are excluded.

\emph{Causal-semantic} contracts require downstream use to preserve responsibility, evidential support, and scope. A missed handoff ignores a delivered supplier change. An unsupported assumption promises shipment after a failed stock lookup. Unsupported reconstruction presents an unreadable invoice's inferred amount as observed. Source scope loss extends one supplier's evidence to all suppliers. Source-view scope loss treats one page as all orders. Evidence status escalation turns ``unverified'' into ``verified.''

\emph{Eventual-semantic} contracts require agreement among designated final states by a declared endpoint. Unresolved final divergence leaves an order both active and canceled. Competing final answers submit conflicting totals without selecting one. An unresolved final evidence state leaves required checks both complete and pending.

\begin{figure}[!tp]
\centering
\includegraphics[width=\linewidth]{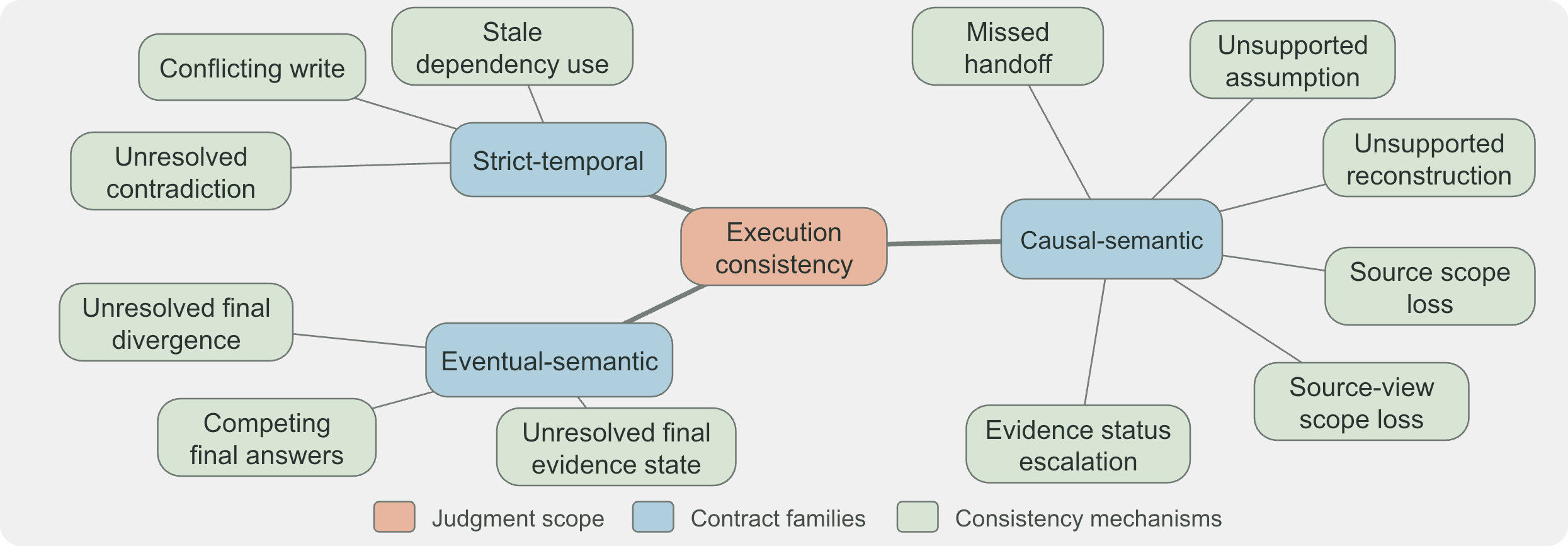}
\caption{Twelve candidate mechanisms grouped by state use, information handoffs, and final-state agreement. Conditions and exclusions: Table~\ref{tab:mechanism_predicates}.}
\label{fig:mechanism_taxonomy}
\end{figure}

Figure~\ref{fig:mechanism_taxonomy} groups twelve candidate mechanisms within these families. Families organize duties; mechanisms identify suspected failures; RQ1's five factors test their evidence requirements (Section~\ref{sec:information_requirements}). An execution can involve multiple families. Appendix~\ref{app:mechanisms} gives detailed conditions and exceptions.

Judging these duties requires more than the updated record. Writing a revocation does not establish receipt; receipt does not establish use. A cancellation request does not establish a successful effect lasting through the deadline. No logged cancellation may mean either that none occurred or that its channel was not recorded.

The execution record $T$ preserves these distinctions by linking messages, memory operations, tool calls and results, handoffs, and final actions. It records event order, who could see a state, which actions used it, and which updates replaced it. Coverage records identify captured channels and times. The recording endpoint says when logging stopped, not whether all relevant events were captured. Appendix~\ref{app:decision_rules} gives the formal schema.

\subsection{Evidence-Based Consistency Judgments}
\label{sec:predicates}

Return to the procurement agent that orders after receiving the revocation. It must cancel by the checkpoint. An authorized cancellation that succeeds on time and remains effective fulfills this duty: no violation (N). Complete relevant records with no qualifying cancellation establish a violation (V). If the cancellation channel was not fully recorded and fulfillment remains unresolved, neither conclusion is justified: the verdict is undetermined (U). U is an evidential outcome, not a confidence score: the record justifies neither fulfillment nor violation. These judgments concern the cancellation duty, not whether the earlier stale use occurred (Figure~\ref{fig:conflict_gap}).

A \emph{finding} $F$ is a suspected problem to check, not a confirmed violation. Given the contract $P$ and execution record $T$, four questions determine its verdict. The first two identify the problem and duty; the last two assess the evidence and response:

\begin{enumerate}
\item \textbf{What might be wrong ($C$)?} Using an old approval is not necessarily a problem. Compare records about the same entity, attribute, and scope (their \emph{semantic key}). Using that approval after a verified replacement became visible to the agent is stale use. Historical references and contract-permitted pinned versions are excluded.

\item \textbf{Who must do what, and by when ($O$)?} A suspected problem does not establish that a response is already due. The contract and activation evidence establish the actor, response, and deadline. An activated duty still binding at its deadline is \emph{active-due}; before then it is pending. Missing, ambiguous, or conflicting policy yields uncertainty, not absence. Established absence of an applicable duty yields N.

\item \textbf{Is the relevant process fully observed ($W$)?} A cancellation may occur without appearing in the log. Check that relevant channels and causal dependencies are recorded, event order is reliable, and state visibility is established. The deadline must be known and recording must extend through it. W is true (a \emph{certified window}) if all six hold, false if any fails, and unknown otherwise.

\item \textbf{Was the duty fulfilled, and did the response remain valid ($D$)?} A cancellation request alone does not establish fulfillment. Check for a response from the responsible or authorized actor, linked to the duty and matching its key and scope. It must occur after activation and by the deadline. Its effect must succeed and remain valid at the deadline or recording endpoint. These conditions define \emph{admissible} fulfillment. Once it is established, the duty is marked \emph{discharged} before judgment (Table~\ref{tab:discharge}).
\end{enumerate}

A valid response can establish fulfillment. Establishing non-fulfillment also requires observing where and when a response could occur. Figure~\ref{fig:verdict_tree} therefore checks established fulfillment before requiring a full observation window. The complete rule table is in Appendix~\ref{app:decision_rules}.

\begingroup
\newcommand{\vttext}[2]{\ifdefined\verdictTreeChinese#2\else#1\fi}
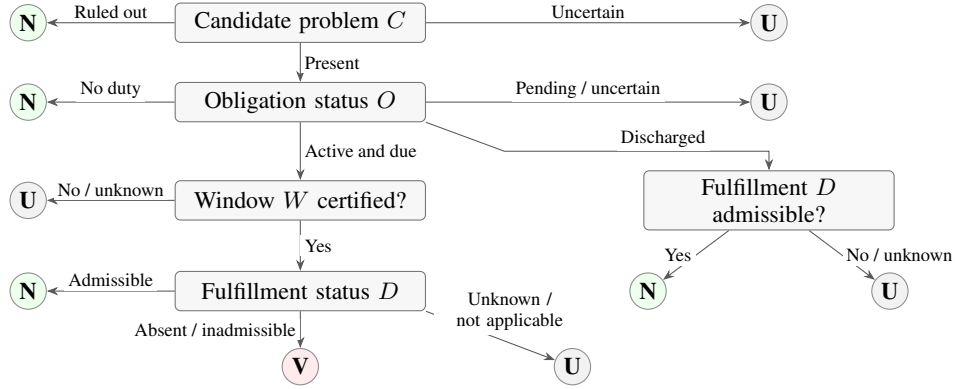
\begin{figure}[!tp]
\centering
\begin{tikzpicture}[
  x=1cm,y=1cm,
  check/.style={draw=black!55,fill=black!3,rounded corners=2pt,
    text width=3.1cm,minimum height=0.52cm,align=center,
    font=\small,inner sep=3pt},
  result/.style={draw=black!45,circle,minimum size=0.48cm,
    inner sep=1pt,font=\small\bfseries},
  branch/.style={-{Stealth[length=1.5mm]},draw=black!65},
  note/.style={font=\scriptsize,fill=white,inner sep=1.5pt,align=center}]
\node[check] (c) at (4.4,0) {\vttext{Candidate problem $C$}{候选问题$C$}};
\node[result,fill=green!7] (cn) at (0.8,0) {N};
\node[result,fill=black!5] (cu) at (10.6,0) {U};
\draw[branch] (c) -- node[note,above]{\vttext{Ruled out}{已排除}} (cn);
\draw[branch] (c) -- node[note,above]{\vttext{Uncertain}{不确定}} (cu);

\node[check] (o) at (4.4,-1.05) {\vttext{Obligation status $O$}{义务状态$O$}};
\draw[branch] (c) -- node[note,right]{\vttext{Present}{已成立}} (o);
\node[result,fill=green!7] (on) at (0.8,-1.05) {N};
\node[result,fill=black!5] (ou) at (10.6,-1.05) {U};
\draw[branch] (o) -- node[note,above]{\vttext{No duty}{无适用义务}} (on);
\draw[branch] (o) -- node[note,above]{\vttext{Pending / uncertain}{未到期／不确定}} (ou);

\node[check] (w) at (4.4,-2.35) {\vttext{Window $W$ certified?}{窗口$W$已认证？}};
\draw[branch] (o) -- node[note,right]{\vttext{Active and due}{活动且已到期}} (w);
\node[result,fill=black!5] (wu) at (0.8,-2.35) {U};
\draw[branch] (w) -- node[note,above]{\vttext{No / unknown}{否／未知}} (wu);

\node[check] (early) at (10.6,-2.35) {\vttext{Fulfillment $D$ admissible?}{履行$D$合格？}};
\draw[branch] (o.south east) -- (7.1,-1.7) -| node[note,pos=0.3,above]{\vttext{Discharged}{已履行}} (early.north);
\node[result,fill=green!7] (en) at (9,-3.55) {N};
\node[result,fill=black!5] (eu) at (12.2,-3.55) {U};
\draw[branch] (early) -- node[note,left]{\vttext{Yes}{是}} (en);
\draw[branch] (early) -- node[note,right]{\vttext{No / unknown}{否／未知}} (eu);

\node[check] (d) at (4.4,-3.55) {\vttext{Fulfillment status $D$}{履行状态$D$}};
\draw[branch] (w) -- node[note,right]{\vttext{Yes}{是}} (d);
\node[result,fill=green!7] (dn) at (0.8,-3.55) {N};
\draw[branch] (d) -- node[note,above]{\vttext{Admissible}{合格}} (dn);
\node[result,fill=red!8] (dv) at (4.4,-4.55) {V};
\draw[branch] (d) -- node[note,left]{\vttext{Absent / inadmissible}{缺失／不合格}} (dv);
\node[result,fill=black!5] (du) at (8,-4.55) {U};
\draw[branch] (d.south east) -- node[note,pos=0.65,above=3pt,text width=1.9cm]{\vttext{Unknown /\\not applicable}{未知／不适用}} (du);
\end{tikzpicture}
\caption{\vttext{Finding-level decisions. N: no violation; V: violation; U: undetermined. The discharged branch permits N without a window certificate.}{判定项的决策树。N：不违规；V：违规；U：无法判定。已履行分支允许在无需窗口认证时判N。}}
\label{fig:verdict_tree}
\end{figure}
\endgroup

\begin{samepage}
In words, V requires an established problem, a binding duty that is due, complete relevant observation, and no response that qualifies as fulfillment:
\begin{equation}
v_F=\mathsf{V}
\quad\Longleftrightarrow\quad
C=\mathsf{present}\ \land\
O=\mathsf{active\_due}\ \land\
W=\mathsf{T}\ \land\
D\in\{\mathsf{absent},\mathsf{inadmissible}\}.
\label{eq:violation}
\end{equation}
\end{samepage}

At trace level, judgments are aggregated under $P$: any confirmed violation yields V. With no confirmed violation, any unresolved applicable finding yields U; otherwise the trace receives N. Timely cancellation fulfills the duty without erasing the earlier stale use. Late cancellation establishes recovery, not timely fulfillment.

\subsection{Information Needed for a Judgment}
\label{sec:consistency_information_necessity}

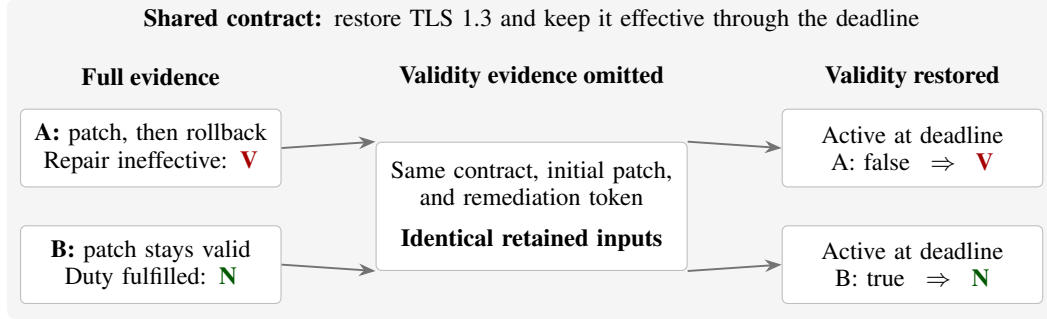
\begin{figure}[!tp]
\centering
\begingroup
\def\pairContract{\textbf{Shared contract:} restore TLS 1.3 and keep it effective through the deadline}
\def\pairFull{Full evidence}
\def\pairRemoved{Validity evidence omitted}
\def\pairIdentical{Identical retained inputs}
\def\pairRestored{Validity restored}
\def\pairA{\textbf{A:} patch, then rollback\\Repair ineffective: \textcolor{red!65!black}{\textbf{V}}}
\def\pairB{\textbf{B:} patch stays valid\\Duty fulfilled: \textcolor{green!35!black}{\textbf{N}}}
\def\pairCommon{Same contract, initial patch,\\and remediation token}
\def\pairFalse{Active at deadline\\A: false $\;\Rightarrow\;$ \textcolor{red!65!black}{\textbf{V}}}
\def\pairTrue{Active at deadline\\B: true $\;\Rightarrow\;$ \textcolor{green!35!black}{\textbf{N}}}
\resizebox{\linewidth}{!}{\begin{tikzpicture}[x=1cm,y=1cm,>=Stealth,
  box/.style={draw=black!25,rounded corners=2pt,fill=white,align=center,inner sep=5pt,font=\small},
  flow/.style={->,draw=black!55,line width=.7pt}]
\path[fill=black!4,rounded corners=3pt] (-.2,.18) rectangle (13.3,4.3);
\node[font=\small,align=center,text width=12.8cm] at (6.55,4.02) {\pairContract};
\node[font=\small\bfseries] at (1.65,3.30) {\pairFull};
\node[font=\small\bfseries] at (6.55,3.30) {\pairRemoved};
\node[font=\small\bfseries] at (11.45,3.30) {\pairRestored};
\node[box,text width=3.0cm,minimum height=1.0cm] (a) at (1.65,2.38) {\pairA};
\node[box,text width=3.0cm,minimum height=1.0cm] (b) at (1.65,.88) {\pairB};
\node[box,text width=3.65cm,minimum height=1.65cm] (q) at (6.55,1.63) {\pairCommon\\[5pt]\textbf{\pairIdentical}};
\node[box,text width=3.0cm,minimum height=1.0cm] (ra) at (11.45,2.38) {\pairFalse};
\node[box,text width=3.0cm,minimum height=1.0cm] (rb) at (11.45,.88) {\pairTrue};
\draw[flow] (a.east) -- (q.north west);
\draw[flow] (b.east) -- (q.south west);
\draw[flow] (q.north east) -- (ra.west);
\draw[flow] (q.south east) -- (rb.west);
\end{tikzpicture}}
\endgroup
\caption{An evaluated pair becomes indistinguishable after removing rollback and validity evidence, then separates after restoration. N/V label the full executions.}
\label{fig:evidence_pair}
\end{figure}

The cancellation example requires a response whose effect lasts. AgentWebBench pair \texttt{pair\_awb\_005} isolates the same requirement for repair: restore TLS~1.3 and maintain it through the deadline (Figure~\ref{fig:evidence_pair}). Both executions record the same patch and remediation token. A rollback invalidates A's repair (V); B's remains effective (N). Keeping only the contract and initial repair leaves identical inputs. Restoring \texttt{active\_at\_deadline}, the evidence-derived field recording lasting validity, separates them again.

These executions need different answers but look identical to the judge: an \emph{input collision}. Formally, sufficiently observed executions have opposite N/V labels but identical retained inputs under the same contract, finding, and deadline. A deterministic judge restricted to these inputs must answer identically, making at least one error. Better reasoning cannot recover the missing distinction. RQ1 tests which evidence preserves it, independently of CAVERT (Section~\ref{sec:information_requirements}; formal criterion in Appendix~\ref{app:decision_rules}).

\section{Applying the Definition with CAVERT}
\label{sec:method}

\begin{figure}[!tp]
\centering
\includegraphics[width=\linewidth,pagebox=cropbox,trim=0bp 12bp 0bp 18bp,clip]{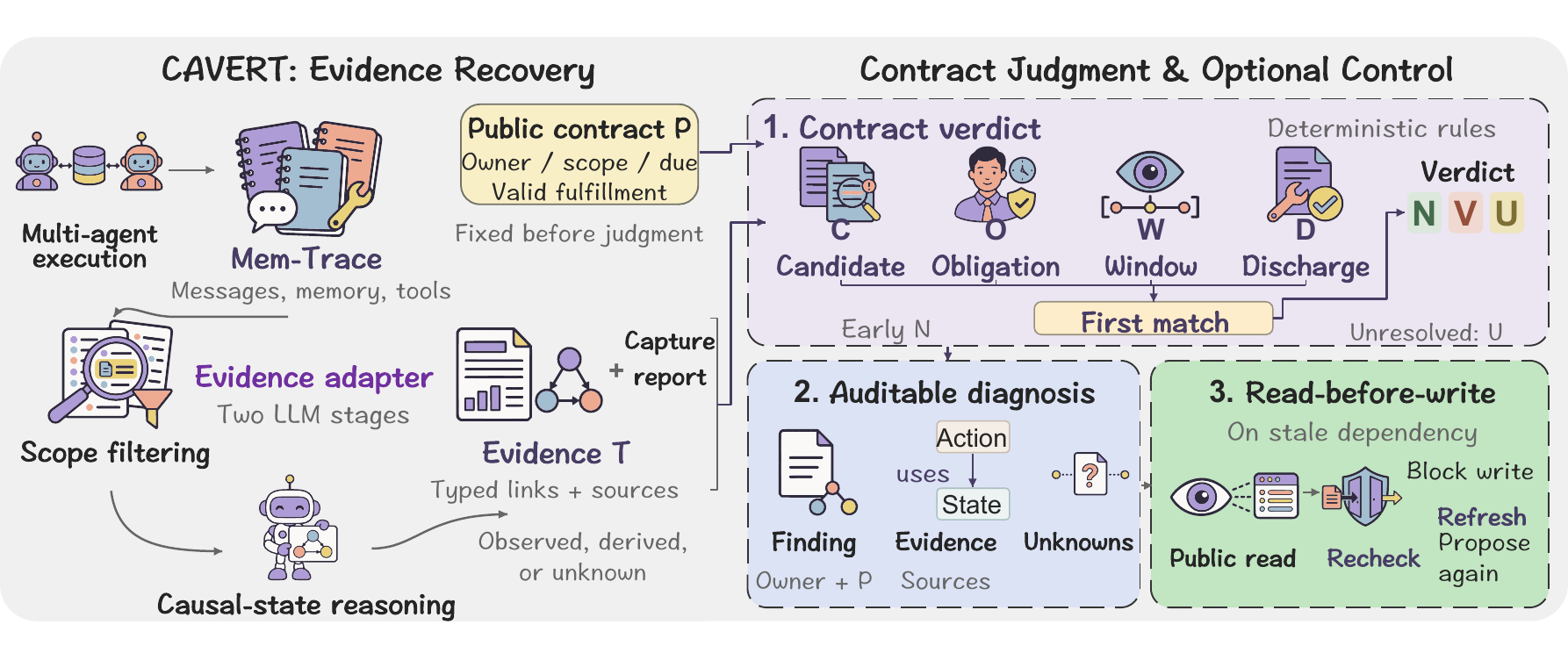}
\caption{CAVERT recovers execution evidence, applies consistency rules, and optionally gates uncommitted writes. The gate refreshes state before a new action proposal (Section~\ref{sec:recovery}).}
\label{fig:framework}
\end{figure}

The definition specifies the evidence required for judgment, but existing logs do not always expose these relationships. CAVERT recovers supported relationships through semantic extraction and source checks, then applies the same deterministic rules. Its findings guide diagnosis, recovery prompts, and pre-commit intervention (Figure~\ref{fig:framework}).

\subsection{From Logs to Check States}
\label{sec:evidence_adapter}

Given a log and contract $P$, two LLM stages recover the evidence needed for judgment. Scope filtering selects relevant messages, memory operations, and tool records. Causal-state reasoning links states to recipients, dependent actions, and response effects.

For procurement, records link the revocation's receipt to the later order's use of the old approval ($C$). The contract supplies the cancellation duty and deadline ($O$). Capture metadata and adapter checks establish coverage ($W$); the cancellation's outcome and lasting effect establish fulfillment ($D$). The adapter retains source-event links and updates the duty's status once fulfillment is established. Unresolved relationships or coverage remain unknown.

LLM interpretation supplies semantic relationships; source checks verify record references (Appendix~\ref{app:extraction_details}). Structured-trace checks read explicit relationship fields (Appendix~\ref{sec:implementation_scope}). Both use the same deterministic verdict rules.

\subsection{Applying the Verdict Rules}
\label{sec:implementation_judgment}

The validator follows Figure~\ref{fig:verdict_tree} without LLM calls and aggregates verdicts as in Section~\ref{sec:predicates}. Model and call budgets appear in Appendix~\ref{app:diagnostic_configuration}.

Each finding reports its verdict, responsible agent, unresolved evidence, and source links (Appendix~\ref{app:schema}). To avoid double-counting a duty, CAVERT groups findings only when they share an obligation, compatible keys and scopes, and a verified connecting path. The most specific upstream cause is primary; downstream consequences are secondary.

\subsection{From Diagnosis to Recovery}
\label{sec:recovery}

The Core gate checks proposed writes before commitment. In procurement, it blocks an order using superseded approval and identifies the state to refresh through a public read before reproposal. Intervention uses the dependency finding before the deadline, without awaiting a trace-level violation.

Supplementary AppWorld recovery blocks writes to a predefined application until a separate read from that application succeeds. State postconditions and official task success are scored afterwards (Appendix~\ref{app:prior_rq3}).

The gate assumes uncommitted intercepted writes, side-effect-free rejection, and an appropriate public read. Committed effects require a separate compensation mechanism.

\section{Experiments}
\label{sec:experiments}\label{sec:evaluation}\label{sec:setup}

Our evaluation separates the definition from its realization. RQ1 tests whether the proposed relationships are necessary for judgment; RQ2 whether CAVERT can recover and apply them in existing logs; RQ3 whether the resulting findings support intervention. Formal checks and contract-query results appear in Appendices~\ref{app:mechanisms} and~\ref{app:e2e3}.

\subsection{RQ1: Validity and Necessity of the Definition}
\label{sec:information_requirements}

\textbf{Execution relationships carry distinctions needed for judgment.} We remove and restore five evidence factors: consumer receipt, action dependency, actor authorization, historical-version eligibility, and repair validity at the deadline. The TLS rollback pair illustrates the last factor: the repair duty stays fixed while evidence of lasting repair is removed and restored (Figure~\ref{fig:evidence_pair}).

The study contains 450 pairs from AgentWebBench, MARBLE, and MECoBench \citep{zhong2026agentwebbench,zhu2025multiagentbench,liu2026mecobench}: 30 per factor per benchmark across 93 task or episode clusters. Each pair shares a contract and query. Blinded human annotation gives 398 opposite-N/V pairs, 38 same-label pairs, and 14 containing U ($\kappa=0.842$). Removal and restoration keep these full-execution labels fixed (Appendix~\ref{app:information_study}).

Among the 398 opposite-N/V pairs, 328 have identical retained inputs after removal (82.4\%), compared with 4 (1.0\%) before removal. Restoration makes 321 of those 328 input pairs distinguishable (97.9\%; Table~\ref{tab:e1_information_necessity}). Task-cluster 95\% intervals are [78.1\%, 86.2\%] and [95.8\%, 99.1\%], respectively. A judge given identical inputs cannot distinguish the opposite labels; restoring the missing relationships recovers that distinction. We next examine whether the defined violations occur in natural executions.

\begin{table}[!tp]
\centering\small
\setlength{\tabcolsep}{3pt}
\caption{Information removal merges oppositely labeled executions.}
\label{tab:e1_information_necessity}
\begin{tabularx}{\linewidth}{@{}Xrr@{}}
\toprule
\textbf{Evidence factor} & \textbf{Removal: merged} & \textbf{Restoration: separated}\\
\midrule
Consumer receipt & 69/80 & 69/69\\
Action dependency & 65/79 & 63/65\\
Actor authorization & 68/81 & 65/68\\
Historical-version eligibility & 57/76 & 55/57\\
Repair validity at deadline & 69/82 & 69/69\\
\midrule
\rowcolor{tableaccent!9}
Total & 328/398 & 321/328\\
\bottomrule
\end{tabularx}
\tabnote{Counts combine three benchmarks. Removal is relative to opposite-label pairs; restoration is relative to merged pairs. Full-view collisions: 4/398. Source breakdowns: Table~\ref{tab:information_full_counts}.}
\end{table}

\textbf{Natural logs reveal recurring execution-consistency problems.} Independent annotation of 360 trajectories (120 each from TraceElephant, AFTraj, and MATM) yields 43 V, 288 N, and 29 U. Among the 43 violating traces, unsupported assumption occurs in 29 (67.4\%) and evidence status escalation in 8 (18.6\%). Both strengthen downstream claims beyond their supporting evidence, giving concrete targets for causal-semantic checks. Multi-label counting gives 47 trace--mechanism associations across six primary mechanisms (Appendix~\ref{app:natural_occurrence}). This distribution describes the sampled sources, including task-failure-only TraceElephant logs.

Together, these studies show that receipt, use, and valid fulfillment carry distinctions that shared content alone does not preserve. This establishes their necessity, but not whether implicit relationships can be recovered reliably from existing logs. RQ2 tests that gap.

\subsection{RQ2: Cross-Source Generalization and Robustness}
\label{sec:diagnostic_evaluation}

\textbf{The evidence requirements support diagnosis from existing logs.} We next apply the criteria to 720 executions from MARBLE, CoffeeBench~\citep{sugiura2026coffeebench}, and MECoBench: four execution models and 60 runs per benchmark--model setting. Diagnostic configurations stay fixed across sources. Generic and Contract LLM use DeepSeek-V4-Flash; CAVERT uses Qwen3.5-Flash filtering, DeepSeek extraction, and deterministic judgment (Appendix~\ref{app:evaluation}).

CAVERT reaches 87.1\% accuracy versus 80.6\% for Contract LLM and 78.9\% for Rules (Table~\ref{tab:diagnosis_summary}). It leads in accuracy and macro-F1 across all 12 settings (Table~\ref{tab:e3_main}), supporting the criteria's use across the tested sources.

\begin{table}[tp]
\centering\small
\caption{Pooled diagnosis on 720 executions; macro-F1 averages N/V/U equally.}
\label{tab:diagnosis_summary}
\begin{tabularx}{\linewidth}{@{}l*{4}{>{\raggedleft\arraybackslash}X}@{}}
\toprule
Metric & Generic LLM & Contract LLM & Rules & \textbf{CAVERT}\\
\midrule
Accuracy (\%) & 64.2 & 80.6 & 78.9 & \textbf{87.1}\\
\rowcolor{tableaccent!9}
Macro-F1 (\%) & 43.5 & 74.8 & 73.8 & \textbf{83.5}\\
\bottomrule
\end{tabularx}
\end{table}

Because call structures differ, we also compare under a common $\$0.0120$ per-execution spending limit. CAVERT exceeds Contract LLM + review by 3.47 pp accuracy and 0.049 macro-F1; both paired 95\% intervals exclude zero. Mean API costs are $\$0.0017$ versus $\$0.0031$ (Appendix~\ref{app:budget_diagnosis}). The advantage persists when the baseline can review and revise within the same spending limit.

\textbf{Format and missing evidence.}
We next test verdict stability under recoding and judgment under reduced evidence. A separate panel gives 180 traces their original view plus four transformations (900 inputs). Recode A renames identifiers and reorders JSON keys; Recode B uses a line-based format with causal pointers. Both preserve relationships and labels. Missing L1 removes delivery confirmations and handshake receipts; L2 also removes pre-deadline invalidations and intermediate action confirmations. Unlike RQ1, reference labels and completeness flags are reassessed from the remaining evidence (Appendix~\ref{app:multiview_diagnosis}).

\begin{figure}[!tp]
\centering
\includegraphics[width=\linewidth,trim=0bp 2bp 0bp 2bp,clip]{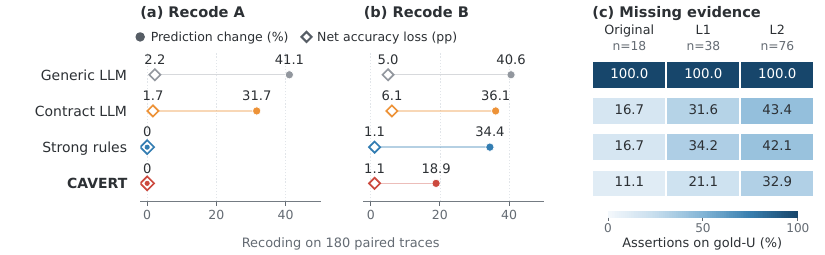}
\caption{Diagnosis under log transformations (180 traces). (a,b) Lines join prediction-change rates and net accuracy losses, not confidence bounds. (c) N/V answer rates on reference-U cases; denominators are shown. Lower is better. Full results: Table~\ref{tab:e3_main}.}
\label{fig:diagnostic_robustness}
\end{figure}

\newcommand{\EThreeOrigAcc}{87.8}
\newcommand{\EThreeRecodeBAcc}{86.7}
\newcommand{\EThreeRecodeBSame}{146}
\newcommand{\EThreeRecodeBAgreement}{81.1}
\newcommand{\EThreeRulesRecodeBSame}{118}
\newcommand{\EThreeRulesRecodeBAcc}{78.9}
\newcommand{\EThreeLtwoAcc}{66.1}
\newcommand{\EThreeLtwoFone}{0.661}
\newcommand{\EThreeRulesLtwoAcc}{57.2}
\newcommand{\EThreeRulesLtwoFone}{0.578}
\newcommand{\EThreeLtwoU}{51}
\newcommand{\EThreeRulesLtwoU}{44}
\newcommand{\EThreeLtwoAssertions}{25}
\newcommand{\EThreeRulesLtwoAssertions}{32}

CAVERT leads all three baselines in accuracy and macro-F1 across all five views (Table~\ref{tab:e3_main}). All its predictions remain unchanged under Recode A; \EThreeRecodeBSame/180 remain unchanged under Recode B, versus \EThreeRulesRecodeBSame/180 for Rules (Figure~\ref{fig:diagnostic_robustness}). The 18 correct-to-wrong and 16 wrong-to-correct transitions yield a 1.1 pp accuracy decline. Individual verdicts remain sensitive to format despite the small net change.

Under Missing L2, CAVERT reaches \EThreeLtwoAcc\% accuracy versus \EThreeRulesLtwoAcc\% for Rules. It nevertheless assigns N to \EThreeLtwoAssertions/76 reference-U and 18/52 reference-V cases: the former are unsupported no-violation judgments; the latter miss violations supported by retained evidence (Appendix~\ref{app:l2_errors}). Diagnosis alone does not establish practical value; RQ3 asks whether these findings improve action under fixed intervention authority.

\subsection{RQ3: Practical Benefits and Intervention Cost}
\label{sec:rq3}
\textbf{Consistency findings guide recovery.} After evaluating diagnosis, we test its use in pre-commit intervention. In procurement, a stale-approval finding directs the agent to refresh the approval state before revising its proposed order. Comparing CAVERT with Rules under the same gate tests the value of this guidance.

\begin{figure}[!tp]
\centering
\includegraphics[width=\linewidth,trim=0bp 1.5bp 0bp 0.5bp,clip]{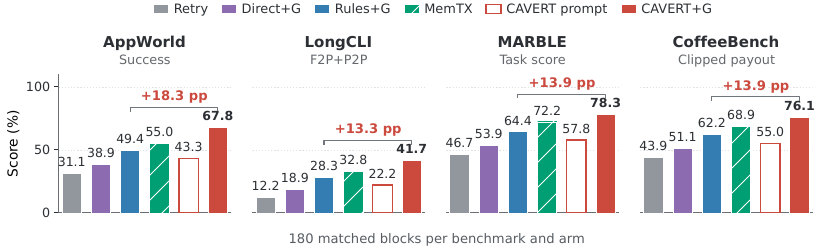}
\caption{Recovery on 180 matched blocks per benchmark. G: shared gate. Gains are CAVERT+G minus Rules+G before rounding (Table~\ref{tab:e4_execution}).}
\label{fig:recovery_outcomes}
\end{figure}

We use 720 matched fault blocks, 180 each from AppWorld~\citep{trivedi2024appworld}, LongCLI~\citep{feng2026longcli}, MARBLE, and CoffeeBench (Figure~\ref{fig:recovery_outcomes}). Within each block, arms share initial states, action budgets, output ceilings, and timeouts. Aggregate scores equally weight benchmark-specific normalized endpoints (Appendix~\ref{app:recovery_config}).

With the same gate, CAVERT improves on Rules+G in all four environments. AppWorld success rises from 89/180 to 122/180 (+18.3 pp), and LongCLI repair without regressions from 51/180 to 75/180 (+13.3 pp). Consistency findings therefore improve recovery with intervention authority held fixed.

We also compare complete systems with their own intervention mechanisms. CAVERT-Gate exceeds MemTX by 8.75 pp in mean normalized score (95\% CI [6.39,11.11]) on the same blocks. All four benchmark intervals are positive (Table~\ref{tab:memtx_recovery_differences}).

\textbf{The benefit of gating depends on diagnostic guidance.} Holding continuation length fixed, we test three diagnostic methods with prompting and with gating. Switching to gating adds 17.7 pp with CAVERT and 8.6 pp with Rules. The difference between these gains is 9.1 pp (95\% CI [4.8,13.4], $p<0.001$; Table~\ref{tab:e5_interaction}). Thus, the same gate is more effective when guided by consistency findings.

Intervention should also preserve normal-task performance. On 480 blocks without injected faults, the CAVERT-minus-Native score difference is $-0.7$ pp (95\% CI [$-2.1$, $+0.7$]). The lower bound exceeds the predeclared $-2.5$ pp loss margin, supporting aggregate non-inferiority (Table~\ref{tab:e6_conditions}).

\textbf{Recovery incurs computational overhead.} On a separate long-horizon cohort, CAVERT+G adds 24.6\% tokens and 5.0 seconds per task over Native. Supplementary AppWorld/LongCLI billing averages $\$0.042$ versus Native's $\$0.030$ per task, including cache discounts (Table~\ref{tab:recovery_overhead}).

RQ1 identifies evidence needed to distinguish fulfillment from violation; RQ2 and RQ3 show how it supports diagnosis and corrective action. Together, the results connect evidence sufficiency to judgment and intervention in multi-agent systems. This progression matters because the same evidence that makes execution behavior judgeable also makes failures actionable before commitment.

\section{Conclusion}
\label{sec:conclusion}
Correct memory is not enough for consistent execution. Identical records can conceal different task-duty outcomes, and no downstream judge can recover distinctions the record failed to preserve. We formalize this gap as execution consistency and identify the evidence needed to judge it. CAVERT shows that these requirements support diagnosis and corrective action. The broader design question is therefore not only whether memory is correct, but whether it preserves enough evidence to judge execution. Reliable multi-agent memory should preserve not only what agents know, but how shared information was received, used, and acted upon.

\label{sec:main_end}
\setlength{\parskip}{6pt}
\section*{Reproducibility Statement}
Appendix~\ref{app:formal} specifies the semantics and evidence schema. Supplementary material provides an implementation snapshot, extraction prompts, an executable example, and 432 constructed executions with reference labels. Paired intervals preserve the stated clustering units. Appendices~\ref{app:evaluation} and~\ref{app:e4e6} specify configurations.

\section*{Ethics Statement}
This work studies execution consistency, diagnosis, and recovery in multi-agent systems using research benchmarks, controlled executions, and model-generated or benchmark-provided traces. We do not collect new human-subject, personal, or sensitive data. Our experiments evaluate task-level execution behavior and system interventions rather than decisions about individuals or demographic groups. Recovery mechanisms operate within the evaluated benchmark environments and are not designed for deployment in consequential real-world settings. We use benchmark data and models for research evaluation and report the evidence requirements, intervention scope, and experimental protocols needed to interpret our results. We follow the ICLR Code of Ethics and aim to support transparent and reproducible evaluation of multi-agent systems.

\section*{AI Use Statement}
Generative-AI tools assisted code, writing, and figure preparation. Reference labels for the information-pair and diagnostic studies were independently human annotated. The authors are responsible for experimental records, references, and final claims.

\bibliographystyle{iclr2027_conference}
\bibliography{iclr2027_conference}

\clearpage
\appendix
\raggedbottom
\setcounter{table}{0}
\renewcommand{\thetable}{A\arabic{table}}
\renewcommand{\theHtable}{appendix.\arabic{table}}
\setcounter{figure}{0}
\renewcommand{\thefigure}{A\arabic{figure}}
\renewcommand{\theHfigure}{appendix.\arabic{figure}}

\appendixpresentation
\renewcommand{\topfraction}{0.93}
\renewcommand{\textfraction}{0.05}
\renewcommand{\floatpagefraction}{0.78}
\setcounter{topnumber}{4}
\setcounter{totalnumber}{5}
\hypertarget{appendix.contents}{}
\pdfbookmark[0]{Appendix Contents}{appendix.guide}
\section*{Appendix Contents}
{\small
\appcontentsgroup{app:formal}{Semantics and Implementation}
\appcontentsentry{app:decision_rules}{Evidence and decision rules}
\appcontentsentry{app:mechanisms}{Mechanism registry}
\appcontentsentry{app:schema}{Trace schema and adapters}
\appcontentsentry{app:policy}{Obligations and effective discharge}
\appcontentsentry{app:distributed_comparison}{Relation to distributed consistency}
\appcontentsentry{app:agent_verification_comparison}{Relation to agent diagnosis and verification}
\appcontentsgroup{app:evaluation}{Shared Evaluation Protocol}
\appcontentsentry{app:annotation_metrics}{Annotation, metrics, and uncertainty}
\appcontentsentry{app:diagnostic_configuration}{Diagnostic and execution configurations}
\appcontentsentry{app:extraction_probe}{Semantic extraction diagnostic probe}
\appcontentsgroup{app:information_study}{Definition Evaluation Details}
\appcontentsentry{app:pair_design}{Paired design and information removal}
\appcontentsentry{app:natural_occurrence}{Mechanism occurrence in natural logs}
\appcontentsgroup{app:e2e3}{Diagnosis and Robustness Details}
\appcontentsentry{app:query_diagnosis}{Contract-query diagnosis (E2)}
\appcontentsentry{app:multiview_diagnosis}{Cross-model and multiview diagnosis (E3)}
\appcontentsentry{app:l2_errors}{Prediction stability and evidence sufficiency}
\appcontentsentry{app:budget_diagnosis}{Diagnosis under a fixed spending limit}
\appcontentsgroup{app:e4e6}{Execution Protocol and Stratification}
\appcontentsentry{app:recovery_config}{Matched blocks and executor configurations}
\appcontentsentry{app:recovery_endpoints}{Recovery controls and outcome definitions}
\appcontentsentry{app:normal_uncertainty}{Normal-task impact and budget sensitivity}
\appcontentsentry{app:prior_rq3}{Supplementary cohorts, outcomes, and costs}
}
\clearpage

\section{Semantics and Implementation}
\label{app:formal}

\FloatBarrier
\subsection{Evidence and Decision Rules}
\label{app:decision_rules}
\subsubsection{Execution Record and Information Loss}

The execution record is
\begin{equation}
T=\langle E,\operatorname{Prec},\operatorname{Causal},
\operatorname{Vis},\operatorname{Sup},P,\operatorname{Capture},H\rangle.
\label{eq:trace}
\end{equation}
Here $E$ denotes events, $\operatorname{Prec}$ reliable precedence, and $\operatorname{Causal}$ typed dependencies. Visibility $\operatorname{Vis}(i,a,j)$ records whether agent $a$ could observe event $i$ before event $j$. Supersession $\operatorname{Sup}$ and contract $P$ determine which states and effects remain eligible for use. $\operatorname{Capture}$ records which channels and segments were captured. The observation horizon $H$ is the point where recording ends. Records are compared using the semantic key $K(e)=\langle\mathrm{entity},\mathrm{attribute},\mathrm{scope}\rangle$.

\begin{appbox}{Retained-record indistinguishability}
Fix a public contract $P$, a target finding $F$, and an evaluation horizon $h$. The recorded horizon $H$ does not guarantee coverage through $h$. Let $q(T)$ retain a specified subset of execution information. Let $\ell_P(F,T,h)$ be the evidence-supported reference judgment. Suppose sufficient evidence establishes opposite labels $\ell_0,\ell_1\in\{\mathsf{N},\mathsf{V}\}$ under the same $(P,F,h)$, but $q(T_0)=q(T_1)$. Any deterministic judge $g$ restricted to that input then satisfies
\begin{equation}
\sum_{j=0}^{1}\mathbf{1}\!\left[g(P,F,h,q(T_j))\ne\ell_j\right]\geq 1.
\label{eq:consistency_projection_collision}
\end{equation}
Identical inputs require identical predictions, matching at most one label. The criterion tests whether a representation preserves verdict-relevant distinctions.
\end{appbox}

\subsubsection{Evidence Domain}

Every primitive queried by the semantics is either observed in the raw execution, reproducibly derived by a versioned adapter rule, or unknown.
Observed and derived values can evaluate to true or false; unknown evidence evaluates to $\mathsf{U}$.
The window certificate uses strong Kleene conjunction to combine these values. Any false requirement makes the certificate false. If none is false but one is unknown, the certificate is unknown. It is true only when all requirements are true.
The finding verdict then follows the ordered finite-trace monitor below.
An uncertified window leaves missing discharge records insufficient to establish non-fulfillment. Earlier decisive rules can still establish N without evaluating the window.

For the truth ordering $\mathsf{F}<\mathsf{U}<\mathsf{T}$,
\[
x\land y=\min(x,y),\qquad
\neg\mathsf{T}=\mathsf{F},\quad
\neg\mathsf{F}=\mathsf{T},\quad
\neg\mathsf{U}=\mathsf{U}.
\]

\subsubsection{Visibility, Reachability, and Frontier}

$\operatorname{VisAt}$ is the relation $\operatorname{Vis}$ in Eq.~\eqref{eq:trace}, evaluated for an agent before an event. True visibility requires one of three grounds: $a$ produced $i$, an active access-control rule includes $a$, or explicit delivery makes $i$ visible before $j$.
An explicit exclusion yields $\mathsf{F}$ only with complete delivery capture through $j$ and no intervening delivery to $a$. Otherwise visibility is unknown.
An empty or missing access-control list is never interpreted as globally visible.
Typed causal reachability is computed over \textsf{data\_dependency}, \textsf{control\_dependency}, \textsf{handoff\_delivery}, \textsf{evidence\_support}, \textsf{supersession}, and \textsf{resolution} edges.
Temporal adjacency or repeated speech by one agent does not create a causal edge.
The typed edges in $\operatorname{Causal}$ record execution relationships; $\operatorname{Sup}$ records which earlier states have retired. They are not interchangeable: a dependency or delivery edge need not retire a state.

A \emph{frontier} contains states still eligible for use: recorded replacements, retractions, or resolutions have not retired them. For trigger $j$ and semantic key $k$, the global pre-trigger frontier is
\begin{equation}
\operatorname{Frontier}_{pre}(j,k)=
\{i\mid i\prec j,\ K(i)\sim k,\ \nexists r:
i\prec r\prec j \land \operatorname{Retires}(r,i)\}.
\end{equation}
Retirement requires an explicit supersession, retraction, version replacement, or effective resolution.
When retirement is unknown, frontier membership is unknown.
The agent-visible frontier uses only the agent's observation history. A retirement changes this frontier only when the agent can see it:
\[
\begin{aligned}
\operatorname{Frontier}_{pre}^{a}(j,k)=\{i\mid {}&i\prec j,\ K(i)\sim k,\ \operatorname{VisAt}(i,a,j)=\mathsf{T},\\
&\nexists r:\ i\prec r\prec j\land\operatorname{Retires}(r,i)\land\operatorname{VisAt}(r,a,j)=\mathsf{T}\}.
\end{aligned}
\]
An unseen retirement does not silently remove a state from $a$'s view.
Unknown visibility or retirement makes any affected membership unknown; downstream predicates may not treat this as established absence.
Checkpoint mechanisms use the frontier named by the public contract. A consumer-specific stale-use finding instead requires the replacement to be visible to that consumer before use.

\subsubsection{Stage Truth Table and Trace Aggregation}

\begin{table}[!tp]
\caption{Ordered stage states and their finding verdicts. Decisive earlier stages short-circuit later checks.}
\label{tab:truth_table}
\label{tab:verdict_summary}
\centering
\footnotesize
\begin{tabularx}{\textwidth}{@{}L{0.13\textwidth}L{0.20\textwidth}L{0.16\textwidth}L{0.21\textwidth}Y@{}}
\toprule
\appendixheader
\textbf{Candidate} & \textbf{Obligation} & \textbf{Window} &
\textbf{Discharge} & \textbf{Final}\\
\midrule
Absent & Not evaluated & Not evaluated & Not applicable & N\\
Unknown & Not evaluated & Not evaluated & Unknown & U\\
Present & Absent & Any & Not applicable & N\\
Present & Pending or uncertain & Any & Any & U\\
Present & Discharged & Any & Admissible & N\\
Present & Discharged & Any & Not admissible or unknown & U\\
Present & Active due & Closed & Admissible & N\\
\rowcolor{tableaccent!8}
Present & Active due & Closed & Inadmissible or absent & \textbf{V}\\
Present & Active due & Closed & Unknown or not applicable & U\\
Present & Active due & Open or unknown & Any & U\\
\bottomrule
\end{tabularx}
\tabnote{N: no violation for this mechanism; V: violation; U: undetermined. ``Any'' means the value cannot change this row's verdict; ``not evaluated'' denotes a short-circuited stage, not an unknown truth value.}
\end{table}

At trace level, one confirmed finding is sufficient for \violation.
If no finding is a violation but any applicable mechanism remains \undetermined, the trace is \undetermined.
The trace is \noviolation\ only when every applicable mechanism has a determined \noviolation\ verdict, including verdicts established by short-circuiting.
Trace aggregation therefore preserves any unknown mechanism that could change the verdict.

A trace-level N concerns the consistency mechanisms applicable under $P$. The recovery gate enforces read-before-write sequencing for uncommitted writes (Section~\ref{sec:recovery}); action authorization remains a separate policy requirement.

\FloatBarrier
\subsection{Mechanism Registry}
\label{app:mechanisms}

Table~\ref{tab:mechanism_predicates} specifies the evidence requirements and exclusions for the twelve mechanisms introduced in Section~\ref{sec:contracts}. A required handoff may bind its recipient when it uses the information, while explicitly permitted alternatives need not be reconciled.

\begin{table}[!tp]
\caption{Required evidence and exclusion conditions for the twelve consistency mechanisms.}
\label{tab:mechanism_predicates}
\centering
\footnotesize
\setlength{\tabcolsep}{4pt}
\begin{tabularx}{\textwidth}{@{}L{0.19\textwidth}L{0.43\textwidth}Y@{}}
\toprule
\appendixheader
\textbf{Mechanism} & \textbf{Required structural evidence} & \textbf{Exclusion or boundary}\\
\midrule
\multicolumn{3}{@{}l}{\textit{Strict-temporal: consistency at a dependency or reconciliation checkpoint}}\\
\addlinespace[3pt]
Conflicting write &
Visible active frontier write; overlapping key/scope; contradictory committed downstream write &
Labelled alternatives or an explicit resolution are not conflicting writes\\
Stale dependency use &
Verified retirement of old state; downstream dependency on it; replacement visible before use &
Explicit version pin, authorized rollback, or historical qualification\\
Unresolved contradiction &
Two contradictory active frontier states at a declared reconciliation checkpoint &
Resolution before the checkpoint prevents this checkpoint candidate\\
\midrule
\multicolumn{3}{@{}l}{\textit{Causal-semantic: preservation of responsibility, support, and evidence scope}}\\
\addlinespace[3pt]
Missed handoff &
Typed handoff with payload, target, delivery condition, and later target dependency action &
Preserve, update, qualify, or resolve before the dependency action\\
Unsupported assumption &
Failed, partial, unknown, or unverified evidence followed by a stronger dependent commitment &
Independent evidence sufficient for the stronger claim\\
Unsupported reconstruction &
Corrupt or unrecoverable source followed by reconstruction presented as observed or verified &
Independent support or an explicitly hypothetical reconstruction\\
Source scope loss &
Source scope is known; downstream scope exceeds it; no support covers the expansion &
Scope-preserving or explicitly narrowed claim\\
Source-view scope loss &
Paged, ranked, filtered, windowed, or top-$k$ view generalized to the global source &
Global completeness is separately established\\
Evidence status escalation &
Failed, partial, unknown, or unverified evidence upgraded to gathered, complete, or verified &
New successful evidence or an auditable verification step\\
\midrule
\multicolumn{3}{@{}l}{\textit{Eventual-semantic: consistency of the effective terminal state}}\\
\addlinespace[3pt]
Unresolved final divergence &
Incompatible task-relevant terminal frontier states remain active at horizon &
Specific upstream mechanism is primary when it explains the same obligation\\
Competing final answers &
Conflicting typed final answers with no canonical selection at horizon &
Policy-authorized labelled alternatives can make the obligation non-binding\\
Unresolved final evidence state &
Terminal evidence availability, completeness, or verification states disagree and affect output &
New evidence, qualification, or resolution before horizon\\
\bottomrule
\end{tabularx}
\end{table}

\FloatBarrier
The $12\times4$ executable suite tests four cases per mechanism: a violation satisfying Eq.~\eqref{eq:violation}, a normal execution without the candidate problem, an authorized and timely discharge, and insufficient evidence requiring \undetermined. Discharge preserves an event-time candidate but can prevent a later checkpoint or terminal candidate. These remain distinct findings even when they concern the same key; grouping follows Section~\ref{sec:implementation_judgment}.

\FloatBarrier
\subsection{Trace Schema and Adapters}
\label{app:schema}

The normalized trace is serialized as \textsf{external-adapted-trace-1.0}.
Each record contains a stable trace identifier, source metadata, a capture manifest, capability evidence, and an ordered event array.
Diagnosis operates on events and their relations, without preselecting a message pair.

For each finding, CAVERT returns
\begin{equation}
y_F=\langle m,\pi,C,O,W,D,v_F,e_r,e_t,\rho\rangle,
\label{eq:diagnosis}
\end{equation}
where $m$ is the mechanism, $\pi$ the responsible principal, $e_r$ the root event, $e_t$ the trigger or due-point event, and $\rho$ a typed dependency path.
Evidence references and abstention reasons accompany the observed path from state break to affected action.
Eq.~\eqref{eq:violation} specifies decisions conditional on these fields; extraction quality is evaluated empirically.

\begin{table}[!tp]
\caption{Public trace fields and their role in evidence construction.}
\label{tab:trace_schema}
\centering
\footnotesize
\begin{tabularx}{\textwidth}{@{}L{0.16\textwidth}L{0.29\textwidth}Y@{}}
\toprule
\appendixheader
\textbf{Object} & \textbf{Required fields} & \textbf{Purpose}\\
\midrule
Source &
\textsf{source\_id}, \textsf{framework}, \textsf{framework\_version}, \textsf{raw\_record\_id} &
Locks provenance and permits source- and framework-stratified evaluation\\
Capture manifest &
\textsf{adapter\_version}, channel states, horizon and truncation states, raw-log reference &
Establishes which negative observations are justified\\
Event &
\textsf{event\_id}, \textsf{sequence}, \textsf{event\_type}, \textsf{actor}, \textsf{content}, \textsf{fields} &
Represents messages, memory operations, tools, retrieval, browser views, handoffs, and finalization\\
Field evidence &
\textsf{status}, \textsf{value}, \textsf{provenance}, \textsf{confidence}, \textsf{reason} &
Separates observed, derived, and unknown inputs and makes derivations auditable\\
State identity &
\textsf{state\_group\_id}, \textsf{state\_dimension}, \textsf{state\_label}, semantic key/scope &
Supports comparisons between heterogeneous sub-states without inventing a global task key\\
Graph evidence &
typed parents, visibility activation, supersession, resolution, and support links &
Supports causal localization, retirement, and discharge-effect checks\\
\bottomrule
\end{tabularx}
\end{table}

Adapters return both the normalized trace and a capability report.
Visibility, semantic keys, obligation force, and causal paths retain their observed or derived support; missing values remain unknown.
Policy comes from explicit task requirements, and observation coverage is checked separately for each finding.
Derived fields store the adapter rule and source event identifiers needed for replay.

\subsubsection{Extraction Instructions and Structured Outputs}
\label{app:extraction_details}

The open-log implementation separates evidence selection, candidate localization, and finding-conditioned review within the selection and reasoning workflow of Section~\ref{sec:evidence_adapter}. These operations can involve multiple page-level calls; model assignments and per-call limits are specified in Appendix~\ref{app:diagnostic_configuration}. The following excerpts are taken from the implemented prompts.

\begin{appbox}{Evidence selection}
\texttt{STATE\_ATOM\_PROMPT} instructs: ``This is exhaustive evidence indexing, not violation judgment'' and ``Do not discard weak, failed, tentative, partial, or contradictory states: record their epistemic status.'' It retains claims, source evidence, shared writes, corrections, handoffs, dependent actions, and final states. Each returned atom has the form \texttt{[event\_id, atom\_type, semantic\_label, epistemic\_status, evidence\_quote]}. Event IDs must come from the supplied page; quotes must be literal source spans, or indexed span IDs resolved by the runtime. An \texttt{overflow} flag requests page splitting when the atom limit is exceeded. Actor identity and source text are restored from the original events.
\end{appbox}

\begin{appbox}{Candidate localization}
\texttt{CROSS\_TRACE\_PAIRING\_PROMPT} compares the indexed events and instructs: ``Do not decide obligation, deadline, discharge, or final violation.'' It lists the twelve registered mechanisms and requires localized root, trigger, and path event IDs. Ordinary task failure or an incorrect answer without a supported execution relationship is insufficient. Its JSON response contains \texttt{recovered\_candidates}; each candidate names a mechanism, a typed key or cited-span identity, and the supporting events. A candidate is a proposal for checking, not a final verdict.
\end{appbox}

\begin{appbox}{Finding-conditioned review}
\texttt{FORMAL\_V4\_TARGETED\_PROMPT} instructs: ``Review the entire supplied page separately for obligation, window, and discharge.'' A public instruction or rule must bind the duty; a normally observed final action must establish its deadline. The prompt further specifies: ``A search retry, acknowledgement, critique, or correction marker without later uptake is not discharge.'' For each finding, it returns one \texttt{finding\_reviews} entry with \texttt{obligation}, \texttt{window}, and \texttt{discharge} assessments. Each assessment contains a status (\texttt{supported}, \texttt{not\_supported\_on\_page}, or \texttt{uncertain}), support event IDs, an initially empty citation list, and a short rationale. The request repeats the public task, completion metadata, and finding anchors; these anchors do not count as evidence that intervening pages were scanned.
\end{appbox}

{\raggedright
The inventory and pairing prompts are defined in \path{experiments/formal_v4_exhaustive_candidate_recovery.py}; the targeted prompt and output normalization are in \path{experiments/formal_v5/semantic_evidence.py}. Their separation prevents the extraction interface from substituting an LLM's final label for the deterministic judgment.\par}

\subsubsection{From Responses to Checked States}
\label{app:extraction_checks}

\paragraph{Normalization and source checks.}
After JSON parsing, \texttt{\_normalize\_response} normalizes candidate identities and path boundaries and applies candidate and predicate JSON schemas. For model-selected event IDs, \texttt{\_hydrate\_citations} copies source text into \texttt{exact\_quote}; the model need not transcribe it. The open-log adapter then checks that the mechanism is registered, root/trigger/support IDs exist, required references are cited, quotes match source records, and path events have valid order. Predicate evidence must match an accepted candidate. Invalid or conflicting items cannot establish the affected state; unresolved evidence is retained rather than treated as absence.

\begin{appbox}{Certifying negative evidence}
\texttt{not\_supported\_on\_page} is a page-local search result, not a false predicate by itself. To establish absent discharge, the targeted verifier records all covered event IDs, page counts, completed reviews, and a deadline event. The adapter checks complete coverage of the supplied record, one negative review per page, and an existing deadline; a positive discharge review prevents a negative certificate. An incomplete or uncertain search cannot establish absent discharge. Scanning all supplied text does not establish that an unrecorded execution channel was captured; capture adequacy and task-specific window evidence remain separate requirements.
\end{appbox}

\paragraph{Field conversion.}
After these checks, the open-log adapter constructs the following states. ``Supported'' below means accepted evidence after normalization and audit, not the unverified JSON assertion alone.

{\small
\begin{tabularx}{\linewidth}{@{}L{0.10\linewidth}Y@{}}
\toprule
\appendixheader
\textbf{Field} & \textbf{Conversion used for judgment}\\
\midrule
$C$ & An accepted, confirmed localized candidate yields \textsf{present}; rejected or unresolved proposals yield \textsf{unknown}. Established absence or refutation is distinguished from missing evidence.\\
$O$ & Supported binding and due obligation yields \textsf{active\_due}; established absence yields \textsf{absent}; unresolved binding yields \textsf{uncertain}.\\
$W$ & Accepted finding-window evidence supplies the window certificate and its supporting event references; open or unresolved windows do not become closed merely because extraction finished.\\
$D$ & Supported admissible fulfillment yields \textsf{admissible}; a certified negative search yields \textsf{absent}; otherwise discharge remains \textsf{unknown}. An active-due duty with admissible fulfillment is marked \textsf{discharged} before judgment.\\
\bottomrule
\end{tabularx}
}

{\raggedright
\texttt{SemanticOpenLogV4Validator} in \path{memory_consistency/v4/semantic_open_log.py} performs the audit and conversion. \texttt{final\_status} in \path{memory_consistency/v4/models.py} applies Table~\ref{tab:verdict_summary}; it makes no LLM call. Source checks verify reference integrity and structural constraints. They do not prove that every proposed semantic relationship, duty interpretation, or window assessment is correct; extraction quality remains an empirical question (Appendix~\ref{app:extraction_probe}).\par}

\subsubsection{Executable Input--Output Example}
\label{app:extraction_example}

This constructed example illustrates the complete candidate and predicate response formats and their deterministic processing. The responses are specified fixtures, not sampled model outputs or additional evaluation observations. The trace is fully observed through a normal final checkpoint, with public completion metadata \texttt{final\_event\_id=e4} and capture metadata permitting negative-evidence search.

\begin{appbox}{Example input: four recorded events}
\begin{tabularx}{\linewidth}{@{}L{0.07\linewidth}L{0.19\linewidth}Y@{}}
\toprule
\appendixheader
\textbf{ID} & \textbf{Actor / event} & \textbf{Recorded content, in order}\\
\midrule
e1 & User / instruction & Use only verified supplier eligibility for the final approval.\\
e2 & Planner / memory write & Supplier eligibility is assumed; verification is unavailable.\\
e3 & Procurement / approval & Approved supplier eligibility based on the assumed status.\\
e4 & Procurement / final checkpoint & Final approval checkpoint reached; the approval remains active.\\
\bottomrule
\end{tabularx}
\end{appbox}

\paragraph{Candidate response.}
The proposed \texttt{unsupported\_assumption} links the assumed state in e2 to its consumption in e3. The cited-span identity uses text present in e2. The complete illustrative response is:

\begin{appcode}{Candidate response \hfill JSON fixture}
{"recovered_candidates": [{
  "mechanism": "unsupported_assumption",
  "semantic_key": null,
  "semantic_identity": {
    "mode": "cited_span", "label": "Supplier eligibility",
    "event_ids": ["e2"]
  },
  "root_event_ids": ["e2"], "trigger_event_ids": ["e3"],
  "causal_path_event_ids": ["e2", "e3"], "citations": []
}]}
\end{appcode}

Normalization retains the selected events and fills their source citations. For example, the e2 citation becomes \texttt{event\_id="e2"} with \texttt{exact\_quote="Supplier eligibility is assumed; verification is unavailable."} It does not add a new semantic assertion.

\noindent\begin{minipage}{\linewidth}
\paragraph{Predicate response.}
The finding-conditioned review selects e1 for the obligation and e4 for the window; the negative discharge assessment covers this example's single page. Its complete illustrative response is:

\begin{appcode}{Predicate response \hfill JSON fixture}
{"finding_reviews": [{
  "mechanism": "unsupported_assumption",
  "root_event_ids": ["e2"], "trigger_event_ids": ["e3"],
  "predicate_reviews": {
    "obligation": {
      "status": "supported", "support_event_ids": ["e1"],
      "citations": [],
      "rationale": "The public instruction requires verified eligibility."
    },
    "window": {
      "status": "supported", "support_event_ids": ["e4"],
      "citations": [],
      "rationale": "Normal completion closes the approval duty."
    },
    "discharge": {
      "status": "not_supported_on_page", "support_event_ids": [],
      "citations": [],
      "rationale": "All events reviewed; no propagated repair occurs."
    }
  }
}]}
\end{appcode}
\end{minipage}

\paragraph{Runtime-generated certificate and verdict.}
The verifier copies source citations for the selected IDs and converts the supported obligation and window to true predicate evidence. It generates the following discharge-search certificate; coverage counts are runtime bookkeeping, not model-provided verdicts:

\begin{appcode}{Discharge-search certificate \hfill Runtime-generated}
{"mechanism": "unsupported_assumption",
 "root_event_ids": ["e2"], "trigger_event_ids": ["e3"],
 "deadline_event_id": "e4",
 "covered_event_ids": ["e1", "e2", "e3", "e4"],
 "page_count": 1, "completed_page_count": 1,
 "search_conclusion": "no_admissible_discharge",
 "page_reviews": [{"page_index": 0,
   "status": "no_admissible_discharge_on_page",
   "rationale": "All events reviewed; no propagated repair occurs."}]}
\end{appcode}

\begin{appbox}{Checked states and resulting verdicts}
Replaying these fixtures through normalization, \texttt{TargetedFindingVerifier}, and \texttt{SemanticOpenLogV4Validator} gives:
\par\smallskip
\begin{tabularx}{\linewidth}{@{}L{0.28\linewidth}Yc@{}}
\toprule
\appendixheader
\textbf{Input condition} & \textbf{Checked states / consequence} & \textbf{Verdict}\\
\midrule
Original fixtures & $C=\mathsf{present}$, $O=\mathsf{active\_due}$, $W=\mathsf{T}$, $D=\mathsf{absent}$ & \textbf{V}\\
Only \texttt{completed\_page\_count} changed to zero & The negative certificate is invalid; $D$ becomes unknown. & \textbf{U}\\
Required candidate quote absent from the source & $C$ becomes unknown. & \textbf{U}\\
\bottomrule
\end{tabularx}
\par\smallskip
These checks demonstrate the conversion and rejection behavior; they do not measure an LLM's ability to produce the illustrated responses.
\end{appbox}

\FloatBarrier
\subsection{Obligations and Effective Discharge}
\label{app:policy}

Contracts are fixed before judgment from explicit task instructions and declared policies. Controlled pairs share a case-level specification; external logs use public task metadata or user-instruction messages. Each obligation records its identifier, key and scope, responsible principals, authorized resolvers, activation, deadline, discharge modes, and scoped exceptions. Execution evidence separately establishes agent visibility, activation, and fulfillment. Due status follows these conditions rather than a coarse policy profile.

\begin{table}[!tp]
\caption{Public obligations and their permitted effects. A: policy binding. B: observable conditions for discharge.}
\label{tab:binding}
\label{tab:discharge}
\centering
\footnotesize
\parbox{\linewidth}{\textbf{A. Binding policy}}\par\smallskip
\footnotesize
\begin{tabularx}{\textwidth}{@{}L{0.25\textwidth}Y@{}}
\toprule
\appendixheader
\textbf{Claim or policy state} & \textbf{Default binding interpretation}\\
\midrule
Hard constraint, verified fact, or evidence state &
\textsf{must\_preserve}; exploratory context does not relax provenance, source scope, supersession, or evidence-status invariants\\
Recommendation &
\textsf{should\_consider} only when policy requires justification, resolution, or no unresolved deviation\\
Hypothesis or brainstorm &
Optional only when exploration is explicit and no invariant is contradicted\\
Labelled alternatives &
Non-binding only when the relaxation names the mechanism/scope and declares that no canonical resolution is required\\
Missing, ambiguous, or conflicting policy &
Obligation is \textsf{uncertain}; no hidden domain threshold is substituted\\
\bottomrule
\end{tabularx}
\par\medskip
\parbox{\linewidth}{\textbf{B. Permitted discharge}}\par\smallskip
\footnotesize
\begin{tabular}{lp{0.72\textwidth}}
\toprule
\appendixheader
\textbf{Mode} & \textbf{Required observable effect}\\
\midrule
\textsf{preserve} & Downstream state retains the obligated payload and evidence strength\\
\textsf{update} & Stronger or newer evidence replaces the old state while retaining provenance\\
\textsf{resolve} & Conflicting branches are explicitly retired and one effective state is established\\
\textsf{qualify} & Normative force or scope is narrowed to what the evidence supports\\
\textsf{authorized\_defer} & Policy permits delay and the event names owner, condition, and a strictly later deadline\\
\bottomrule
\end{tabular}
\tabnote{All modes require an authorized actor, the same key and scope, a typed causal path, and an effect after activation and no later than the deadline. The effect must remain active at the evaluation frontier.}
\end{table}

An obligation is \textsf{absent} when public policy establishes that no relevant responsibility exists.
It is \textsf{pending} before its deadline.
It is \textsf{uncertain} when a binding fact is unknown.
The adapter assigns \textsf{active\_due} once grounding, target responsibility, activation, and due status are established. Confirming an admissible discharge changes that state to \textsf{discharged}.
``Not due'' therefore means pending, not absent.
The adapter updates $O$ before verdict evaluation. An established due obligation with admissible $D$ is supplied as $O=\mathsf{discharged}$. The final status function reads $C,O,W,D$ without rewriting $O$. Independently supplied states may instead contain $O=\mathsf{active\_due},D=\mathsf{admissible}$ (Table~\ref{tab:truth_table}). This combination requires a closed W. The two rows thus represent different inputs, not a choice of rules for the same normalized evidence. For a present candidate and unknown W, the discharged input yields N; the active-due input yields U.

Later events can revoke a discharge; the permitted effect must still hold at the deadline or horizon (Table~\ref{tab:discharge}B).

\begin{appbox}{Early N and persistence evidence}
\phantomsection
\label{app:early_n}
``Early'' denotes rule order, not a judgment made before future effects are known. For a present candidate, $O=\mathsf{discharged}$ and $D=\mathsf{admissible}$ establish N without requiring the full W certificate. Admissibility still requires evidence of authorization, matching key and scope, causal connection, timing, and continued effect through the evaluation point. A successful repair receipt alone does not establish persistence if subsequent capture is incomplete.

For example, an agent receives a replacement at event 1, uses the old state at event 2, and repairs it at event 3 before deadline 9. A lasting authorized repair with complete relevant subsequent records establishes discharge. Explicit audience records can establish local visibility while global coverage, and hence W, remains unknown. If persistence is unresolved, D is unknown and this branch yields U. If the repair is retired by event 9, no alternative response exists, and W is closed, the verdict is V.
\end{appbox}

\subsubsection{Evidence Paths and Evaluation Roles}
\label{sec:implementation_scope}
The formal-v4 \texttt{final\_status} function applies Table~\ref{tab:verdict_summary} to $C,O,W,D$. E2/E3 recover these states through two LLM stages and score the complete log-to-verdict path (Appendix~\ref{app:diagnostic_configuration}).

Structured-trace checks read explicit fields instead. Audiences or completed typed delivery establish visibility; supersession and effect-retirement links track response persistence. Persistence requires a known horizon, no truncation, and complete order, causal, memory, tool, and message records. Global visibility coverage is checked separately by W. Missing required links or persistence evidence leave D unknown.

\FloatBarrier
\subsection{Relation to Distributed Consistency}
\label{app:distributed_comparison}

Linearizability concerns an object's sequential specification \citep{herlihy1990linearizability}. Two executions can share the same approval-register history yet differ in whether procurement cancels an order after receiving a revocation. A specification covering order actions can enforce that duty. Our judgment concerns its fulfillment, beyond the register history alone.

COPS tracks dependencies before exposing updates \citep{lloyd2011cops}. Our causal-semantic contracts additionally check meaning: a correctly ordered handoff can still turn ``filtered records'' into ``all records.'' Likewise, eventual convergence \citep{vogels2009eventually} differs from resolving designated task states by a specified deadline. These distinctions concern the specification and evidence, rather than a stronger consistency model over the same object.

Application invariants and semantic checks already support coordination, enforcement, and repair \citep{bailis2014coordination,balegas2015indigo,chang2025sagallm}. Commitment protocols and runtime verification supply obligation lifecycles and three-valued judgments \citep{singh1999commitments,mallya2006commitment,fornara2011obligations,bauer2011runtime}. We specify the evidence linking generated handoffs and tool outcomes to these duties, and test its role through controlled pairs. U covers unresolved execution evidence or policy binding.

\FloatBarrier
\subsection{Relation to Agent Diagnosis and Verification}
\label{app:agent_verification_comparison}

DoVer tests failure hypotheses through interventions, while AgenTracer attributes failures to agents and steps \citep{ma2026dover,zhang2026agentracer}. VIRF checks plans against safety constraints; VeriCoT checks formalized reasoning; ST-WebAgentBench evaluates policy-compliant task completion \citep{wu2026virf,feng2026vericot,levy2026stwebagentbench}. CAVERT judges task-obligation fulfillment from visibility, dependencies, response effects, and coverage.

\FloatBarrier
\section{Shared Evaluation Protocol}
\label{app:evaluation}\label{app:baselines}

\FloatBarrier
\subsection{Annotation, Metrics, and Uncertainty}
\label{app:annotation_metrics}

E1 varies information under a fixed public contract. E2 scores queries registered before method predictions. E3 preserves original labels for lossless views and judges missing-evidence views from what remains. E1 uses two independent annotators with agreement $\kappa=0.842$ and third-party adjudication of disagreements. The development sets contain 30 pairs for E1 and 90 traces for E2 and do not enter formal denominators.

\begin{appbox}{Metric definitions}
\textbf{Three-way macro-F1} averages the N, V, and U classwise F1 scores equally; zero-denominator classes receive zero.
\par\smallskip
\textbf{Decision coverage (Cov.)} is the fraction of all units answered N or V, including reference-U units. It measures definite-answer frequency rather than log completeness.
\par\smallskip
\textbf{Binary violation F1} excludes reference U: a predicted U on V is a false negative, while a predicted U on N is not a false positive.
\end{appbox}

Task-cluster resampling keeps each task's arms, queries, models, seeds, and views together. E1 uses 10,000 percentile-bootstrap replicates; all-zero or all-one cells yield degenerate empirical intervals.

\FloatBarrier
\subsection{Diagnostic and Execution Configurations}
\label{app:diagnostic_configuration}

B3/B4 both disable thinking, with different output ceilings (Table~\ref{tab:delivery_config}). Rules (also labeled Strong rules) uses \texttt{PublicContractRules} to evaluate public-contract evidence deterministically and propagate unknown required evidence to U. AgentRx uses a contract resolver because failure attribution differs from N/V/U judgment. CAVERT's two LLM stages supply scope and causal-state reasoning; the final status function maps the recovered check states to N/V/U deterministically.

\begin{table}[!tp]
\centering\footnotesize
\caption{E2--E3 diagnostic and execution configurations.}
\label{tab:delivery_config}
\begin{tabularx}{\linewidth}{@{}lYY@{}}
\toprule
\appendixheader
Component & Model / engine & Inference configuration\\
\midrule
B3 & deepseek-ai/DeepSeek-V4-Flash-2026-07 & $T=0$, top-$p=1$, 2,048 output tokens; thinking disabled\\
B4 & deepseek-ai/DeepSeek-V4-Flash-2026-07 & $T=0$, top-$p=1$, 4,096 output tokens; thinking disabled\\
Rules & \texttt{PublicContractRules} & Deterministic public-contract checks; unknown required evidence propagates to U\\
AgentRx & AgentRx-Magentic-One-v4.0 + adapter & Adapter: agentrx-magentic-v4-loss-aware-1.0\\
CAVERT & S1: Qwen3.5-Flash; S2: DeepSeek-V4-Flash & $T=0$, top-$p=1$, 4,096 output tokens per model call in each stage; scope filtering then causal-state inference\\
Execution agent & deepseek-ai/DeepSeek-V4-Flash-2026-07 & $T=0.2$, top-$p=0.95$, 4,096 output tokens; ReAct loop\\
\bottomrule
\end{tabularx}
\tabnote{B3/B4 use direct and contract prompts. Diagnostic configurations share public inputs and use the method-specific output ceilings above; CAVERT's ceiling applies per call. These are limits rather than measured consumption. Core E4 executor settings appear in Appendix~\ref{app:e4e6}.}
\end{table}

\paragraph{Stage interfaces.}
Given the log and contract, Qwen3.5-Flash selects relevant messages, memory operations, and tool records. DeepSeek-V4-Flash links this evidence to recipients, dependent actions, and effects. The adapter constructs $T$ and finding-level $C,O,W,D$ states, retaining supporting record references and leaving unresolved fields unknown. The validator then applies Table~\ref{tab:verdict_summary} and aggregates findings without further LLM calls. The 4,096-token output ceiling applies to each extraction call, not to the full two-stage process.

The execution-table labels DeepSeek, Gemini, Qwen, and Llama denote DeepSeek-V4-Flash, Gemini-2.5-Flash, Qwen3.8-27B-FP8, and Llama-3.3-70B-Instruct-FP8, respectively. These are execution models; Qwen3.5-Flash is used in extraction. The four-model diagnosis and supplementary recovery panels are distinct from Core E4's two executor configurations, specified in Appendix~\ref{app:e4e6}.

\FloatBarrier
\subsection{Semantic Extraction Diagnostic Probe}
\label{app:extraction_probe}

The Oracle probe tests evidence quality under fixed verdict rules on a separate calibration set of 432 traces from 144 task clusters (Table~\ref{tab:ablation_oracle}). It replaces CAVERT's two-stage LLM extraction with reference semantic inputs. The structural baseline changes both extraction and adjudication.

\begin{table}[!tp]
\centering\footnotesize
\caption{System configurations and a reference-semantic-input probe ($N=432$ traces; 144 task clusters).}
\label{tab:ablation_oracle}
\begin{tabularx}{\linewidth}{@{}>{\raggedright\arraybackslash}X>{\raggedright\arraybackslash}X>{\raggedright\arraybackslash}Xrr>{\raggedright\arraybackslash}X@{}}
\toprule
\appendixheader
Configuration & Evidence Source & Adjudication Rules & 3-Way Acc & Error Count & Violation Prec / Rec / F1\\
\midrule
1. Structural Baseline & Heuristic (No LLM) & Heuristic Rules & 58.3\% & 180 errors & 45.8\% / 64.7\% / 0.537 (FP=117)\\
2. End-to-End CAVERT & 2-stage LLM & Deterministic Rules & \textbf{83.3\%} & \textbf{72 errors} & \textbf{91.7\% / 64.7\% / 0.759 (FP=9)}\\
3. Oracle + Rules & Reference semantics & Deterministic Rules & \textbf{97.9\%} & \textbf{9 errors} & \textbf{100.0\% / 100.0\% / 1.000 (FP=0)}\\
\bottomrule
\end{tabularx}
\tabnote{The structural baseline changes both extraction and adjudication, making it a full-system comparison. CAVERT reduces FP from 117 to 9, while recall remains 64.7\% (TP=99, FN=54). The non-deployable Oracle probe has 9 errors versus 72 end-to-end errors: a net reduction of 63 (87.5\%). Its remaining errors are Gold N predicted as U; violation FP/FN are zero on this set. With rows and columns ordered V/N/U, the Oracle matrix is $((153,0,0),(0,171,9),(0,0,99))$, giving macro-F1 0.977.}
\end{table}

Reference semantic inputs raise accuracy from 83.3\% to 97.9\% and reduce errors from 72 to 9 (87.5\%); the remaining errors are N cases predicted U. RQ1 identifies relationships that preserve verdict distinctions; this probe shows the value of supplying them accurately. Together, they motivate recording these relationships during memory use, message delivery, and tool execution, so future checkers can read them directly. CAVERT provides the complementary log-based route for existing systems.

\FloatBarrier
\section{Definition Evaluation Details}
\label{app:information_study}

\FloatBarrier
\subsection{Paired Design and Information Removal}
\label{app:pair_design}

The three sources each contribute 150 pairs across five factors. The official query and public contract are the same within a pair. Consumer receipt and version eligibility matter only where the contract makes them relevant; an unreceived update is not by itself proof of a violation. A repair must remain admissible at the deadline, rather than merely have appeared earlier. Full, reduced, and selectively restored representations are canonicalized while preserving semantically relevant identity and ordering. Here, \emph{full} denotes the normalized representation before ablation, not the complete execution record. Equality is tested on this representation; reference labels assess duty fulfillment in the underlying executions under the shared contract.

Reference judgments assess duty fulfillment using the pre-removal evidence and remain fixed across views. Each ablation removes the selected relationship; restoration adds back its verdict-relevant information from the full view. For example, \texttt{active\_at\_deadline} in Figure~\ref{fig:evidence_pair} summarizes whether the repair remains valid at the deadline after accounting for later events.

The 398 opposite-label pairs contain 4 full-view collisions, 328 reduced-view collisions, and 321 restored separations. Every reported factor cell satisfies $S\le C_{\rm minus}-C_{\rm full}$. This bound assumes reduction is a deterministic function of the full representation and restoration introduces no information outside that full view.

Table~\ref{tab:information_full_counts} reports all benchmark--factor cells, each containing 30 pairs. The 38 same-label pairs and 14 pairs containing U remain in the collected sample but are excluded from collision denominators.

\begin{table}[!tp]
\centering\small
\caption{Full sample accounting and representation collisions by information factor.}
\label{tab:information_full_counts}
\begin{tabularx}{\linewidth}{@{}l*{4}{>{\raggedleft\arraybackslash}X}@{}}
\toprule
\appendixheader
Information factor & $D/A$ & $C_{\rm full}/D$ & $C_{\rm minus}/D$ & $S/C_{\rm minus}$\\
\midrule
\multicolumn{5}{l}{\textit{AgentWebBench}}\\
Consumer receipt & 27/30 & 0/27 & 23/27 & 23/23\\
Action dependency & 26/30 & 0/26 & 21/26 & 21/21\\
Actor authorization & 28/30 & 1/28 & 24/28 & 23/24\\
Historical-version eligibility & 25/30 & 0/25 & 19/25 & 18/19\\
Repair validity at deadline & 27/30 & 0/27 & 22/27 & 22/22\\
Subtotal & 133/150 & 1/133 & 109/133 & 107/109\\
\midrule
\multicolumn{5}{l}{\textit{MARBLE}}\\
Consumer receipt & 27/30 & 0/27 & 24/27 & 24/24\\
Action dependency & 27/30 & 1/27 & 23/27 & 22/23\\
Actor authorization & 26/30 & 1/26 & 22/26 & 21/22\\
Historical-version eligibility & 26/30 & 0/26 & 20/26 & 19/20\\
Repair validity at deadline & 28/30 & 0/28 & 24/28 & 24/24\\
Subtotal & 134/150 & 2/134 & 113/134 & 110/113\\
\midrule
\multicolumn{5}{l}{\textit{MECoBench}}\\
Consumer receipt & 26/30 & 0/26 & 22/26 & 22/22\\
Action dependency & 26/30 & 1/26 & 21/26 & 20/21\\
Actor authorization & 27/30 & 0/27 & 22/27 & 21/22\\
Historical-version eligibility & 25/30 & 0/25 & 18/25 & 18/18\\
Repair validity at deadline & 27/30 & 0/27 & 23/27 & 23/23\\
Subtotal & 131/150 & 1/131 & 106/131 & 104/106\\
\midrule
\rowcolor{tableaccent!9}
Total & 398/450 & 4/398 & 328/398 & 321/328\\
\bottomrule
\end{tabularx}
\tabnote{$A$: collected pairs; $D$: opposite N/V labels; $S$: reduced collisions separated by restoring information. The 38 same-label and 14 U-containing pairs remain in sample accounting but not in $D$.}
\end{table}

\FloatBarrier
\subsection{Mechanism Occurrence in Natural Logs}
\label{app:natural_occurrence}

The natural-log annotation release, frozen on September 2, 2026, contains 360 traces: 120 each from TraceElephant \citep{chen2026seeingelephantbenchmarkfailure}, AFTraj \citep{zhang2026agentforesight}, and MATM \citep{kim2026multiagenttransactivememory}. Two annotators independently assess the twelve mechanisms without seeing CAVERT predictions; a third adjudicates disagreements. The separate 30-trace calibration set is excluded.

TraceElephant contributes 40 Captain-Agent and 80 Magentic-One task-failure traces. AFTraj contributes 90 source-labeled safe and 30 diagnosed natural-failure traces, sampled within domains. MATM contributes one randomly selected population run per sampled task: 40 ALFWorld and 80 WebArena tasks. Injected AFTraj traces, MATM expert demonstrations, and single-agent SWE-Agent traces are excluded. Source labels and task outcomes are hidden during annotation. The source strata define the descriptive sample; its proportions characterize these selected logs rather than deployment-wide prevalence.

\begin{table}[!tp]
\centering\small
\caption{Reference verdicts in the 360-trace natural-log panel.}
\label{tab:natural_sources}
\begin{tabularx}{\linewidth}{@{}lXrrrr@{}}
\toprule
\appendixheader
Source & Sampling unit and stratum & Traces & V & N & U\\
\midrule
TraceElephant & Released MAS task-failure traces & 120 & 38 & 82 & 0\\
AFTraj & Safe / diagnosed natural-failure traces & 120 & 2 & 105 & 13\\
MATM & One population run per sampled task & 120 & 3 & 101 & 16\\
\midrule
\rowcolor{tableaccent!9}
Total & & 360 & 43 & 288 & 29\\
\bottomrule
\end{tabularx}
\tabnote{V/N/U are consistency reference labels, not task-success labels or detector predictions. Each source has its own sampling frame.}
\end{table}

Table~\ref{tab:natural_mechanisms} counts a trace once for a mechanism when that mechanism's final reference verdict is V. Secondary manifestations are excluded from these primary-mechanism counts. Four violating traces have two primary mechanisms, yielding 47 trace--mechanism associations across 43 violating traces. Percentages therefore use 43 as the denominator and need not sum to 100\%.

\begin{table}[!tp]
\centering\small
\caption{Primary violation mechanisms in natural logs.}
\label{tab:natural_mechanisms}
\begin{tabularx}{\linewidth}{@{}Xrrrrr@{}}
\toprule
\appendixheader
Mechanism & TE & AF & MT & Total & \% of 43 V\\
\midrule
\multicolumn{6}{@{}l}{\textit{Strict-temporal}}\\
Conflicting write & 0 & 0 & 0 & 0 & 0.0\\
Stale dependency use & 0 & 0 & 0 & 0 & 0.0\\
Unresolved contradiction & 0 & 0 & 0 & 0 & 0.0\\
\midrule
\multicolumn{6}{@{}l}{\textit{Causal-semantic}}\\
Missed handoff & 0 & 0 & 0 & 0 & 0.0\\
\rowcolor{tableaccent!9}
Unsupported assumption & 24 & 2 & 3 & \textbf{29} & \textbf{67.4}\\
Unsupported reconstruction & 1 & 0 & 0 & 1 & 2.3\\
Source scope loss & 3 & 0 & 0 & 3 & 7.0\\
Source-view scope loss & 4 & 0 & 0 & 4 & 9.3\\
Evidence status escalation & 8 & 0 & 0 & 8 & 18.6\\
\midrule
\multicolumn{6}{@{}l}{\textit{Eventual-semantic}}\\
Unresolved final divergence & 0 & 0 & 0 & 0 & 0.0\\
Competing final answers & 2 & 0 & 0 & 2 & 4.7\\
Unresolved final evidence state & 0 & 0 & 0 & 0 & 0.0\\
\midrule
Trace--mechanism associations & 42 & 2 & 3 & 47 & \\
\bottomrule
\end{tabularx}
\tabnote{TE: TraceElephant; AF: AFTraj; MT: MATM. Counts are distinct traces per primary mechanism with a V reference verdict. A trace can contribute to multiple rows. Secondary labels are excluded; zero means no primary V label in this sample.}
\end{table}

Unsupported assumption is the most frequent primary mechanism: 24 TraceElephant, 2 AFTraj, and 3 MATM traces. Evidence status escalation is next, with 8 TraceElephant traces. Both concern downstream claims that exceed their supporting evidence. Unresolved final evidence state also appears as a secondary manifestation in 25 traces; its zero primary count reflects this distinction. The other zero entries denote no primary V label in this sample. Together, the annotations show that the definition identifies recurring execution problems in natural logs while distinguishing them from task failure alone.

\FloatBarrier
\section{Diagnosis and Robustness Details}
\label{app:e2e3}\label{app:external}

\FloatBarrier
\subsection{Contract-Query Diagnosis (E2)}
\label{app:query_diagnosis}

The supplementary E2 panel evaluates diagnosis at the contract-query level: 930 preregistered queries on 360 executions, 120 each from MARBLE, CoffeeBench, and MECoBench. Query counts are 348, 312, and 270; pooled labels are 425 V, 415 N, and 90 U. Our AgentRx adaptation~\citep{barke2026agentrx} uses a contract resolver to map failure attribution to N/V/U judgments.

\begin{table}[!tp]
\centering\footnotesize
\caption{E2: Comparison over 930 contract queries across three benchmarks.}
\label{tab:e2_diagnosis}
\begin{tabular*}{\linewidth}{@{\extracolsep{\fill}}lrrrrl@{}}
\toprule
\appendixheader
Method & MARBLE & CoffeeBench & MECoBench & Pooled & $\Delta$ [95\% CI]\\
\midrule
Always N & 0.205 & 0.204 & 0.207 & 0.206 & $-0.545\;[-0.578,-0.512]$\\
Generic LLM & 0.465 & 0.463 & 0.470 & 0.466 & $-0.285\;[-0.318,-0.252]$\\
AgentRx + resolver & 0.591 & 0.591 & 0.591 & 0.591 & $-0.160\;[-0.191,-0.130]$\\
Strong rules & 0.632 & 0.631 & 0.622 & 0.629 & $-0.122\;[-0.150,-0.094]$\\
Contract LLM & 0.663 & 0.660 & 0.657 & 0.660 & $-0.091\;[-0.117,-0.064]$\\
CAVERT & 0.753 & 0.747 & 0.753 & 0.751 & Reference\\
\bottomrule
\end{tabular*}
\tabnote{All score columns are three-way macro-F1; benchmark query counts are 348, 312, and 270. Pooled uses the combined matrix. Differences are method minus CAVERT pooled macro-F1, with paired 95\% intervals. Accuracy, violation F1, and coverage are in the stratified appendix table. AgentRx includes a contract resolver.}
\end{table}

\begin{table}[!tp]
\centering\small
\caption{E2: Diagnostic performance by benchmark.}
\label{tab:e2_strata}
\begin{tabular*}{\linewidth}{@{\extracolsep{\fill}}lrrrr@{}}
\toprule
\appendixheader
Method & Macro-F1 & Binary V F1 & Acc. (\%) & Cov. (\%)\\
\midrule
\multicolumn{5}{@{}l}{\textbf{MARBLE ($N=348$)}}\\
Always N & 0.205 & 0.000 & 44.5 & 100.0\\
Generic LLM & 0.465 & 0.749 & 66.4 & 100.0\\
AgentRx + resolver & 0.591 & 0.630 & 64.9 & 90.5\\
Strong rules & 0.632 & 0.714 & 70.1 & 87.1\\
Contract LLM & 0.663 & 0.731 & 72.7 & 91.1\\
CAVERT & 0.753 & 0.783 & 78.2 & 91.1\\
\addlinespace
\multicolumn{5}{@{}l}{\textbf{CoffeeBench ($N=312$)}}\\
Always N & 0.204 & 0.000 & 44.2 & 100.0\\
Generic LLM & 0.463 & 0.750 & 66.0 & 100.0\\
AgentRx + resolver & 0.591 & 0.629 & 64.7 & 90.4\\
Strong rules & 0.631 & 0.709 & 69.6 & 86.9\\
Contract LLM & 0.660 & 0.727 & 72.1 & 90.7\\
CAVERT & 0.747 & 0.770 & 77.2 & 90.4\\
\addlinespace
\multicolumn{5}{@{}l}{\textbf{MECoBench ($N=270$)}}\\
Always N & 0.207 & 0.000 & 45.2 & 100.0\\
Generic LLM & 0.470 & 0.752 & 67.4 & 100.0\\
AgentRx + resolver & 0.591 & 0.620 & 65.2 & 90.7\\
Strong rules & 0.622 & 0.715 & 70.4 & 88.1\\
Contract LLM & 0.657 & 0.730 & 73.0 & 91.5\\
CAVERT & 0.753 & 0.776 & 78.1 & 91.5\\
\addlinespace
\bottomrule
\end{tabular*}
\tabnote{Macro-F1 averages N/V/U equally. Binary V F1 excludes reference-U cases; a predicted U on V is a miss. Cov. is the percentage of all queries answered N or V.}
\end{table}

\begin{appbox}{E2: CAVERT confusion matrices}
Rows are reference labels and columns are predictions, both ordered V,N,U. CAVERT's three benchmark matrices are
\[
M_{\rm MB}=\begin{pmatrix}108&46&5\\9&142&4\\0&12&22\end{pmatrix},\quad
M_{\rm CB}=\begin{pmatrix}94&43&5\\8&126&4\\0&11&21\end{pmatrix},\quad
M_{\rm ME}=\begin{pmatrix}83&37&4\\7&112&3\\0&8&16\end{pmatrix}.
\]
\end{appbox}
CAVERT leads in macro-F1; the generic judge has higher violation recall. Table~\ref{tab:e2_strata} reports per-benchmark metrics. Table~\ref{tab:e2_diagnosis} pools confusion matrices and reports method-minus-CAVERT differences with intervals from matched query predictions.

\FloatBarrier
\subsection{Cross-Model and Multiview Diagnosis (E3)}
\label{app:multiview_diagnosis}

The multiview evaluation contains 180 traces, four methods, and five views (3,600 predictions). Accuracy, macro-F1, coverage, and prediction transitions use the same reference labels and predictions, matched by benchmark, executor, and trace within each view. AgentRx is evaluated in the query-level setting; the multiview panel compares CAVERT with Generic LLM, Contract LLM, and Rules.

\paragraph{Transformations and Evaluation Protocol.}
Original retains the supplied execution log. Recode A consistently renames \texttt{event\_id} and \texttt{actor\_id}, updates their references, and reverses the alphabetical order of JSON keys. Recode B flattens the JSON hierarchy into a line-based log format with explicit causal pointers. Both preserve the recorded relationships and original reference labels.

Missing L1 removes direct prerequisite-delivery confirmations (\texttt{context\_deliver}) and explicit handshake receipts, and sets \texttt{channel\_complete} to false. Missing L2 starts from L1 and also removes pre-deadline state-retirement events (\texttt{effect\_retirement}) and intermediate action confirmations. It additionally sets \texttt{causal\_complete} and \texttt{complete\_through\_deadline} to false. These flags describe lost coverage, rather than evidence that an unrecorded event did not occur. Each removal view receives a reference judgment based on its remaining evidence.

The link to the verdict rules is finding-specific. Each obligation is checked over its relevant channels and time interval. A false required coverage condition prevents that finding from establishing V. An earlier N rule may still apply; otherwise the judgment is U (Table~\ref{tab:verdict_summary}). At trace level, V requires at least one finding with a certified window and a violated duty, even if other findings remain unresolved.

Methods are scored on the same instances and reference labels within each view, using the configurations in Table~\ref{tab:delivery_config}. Cross-view agreement measures prediction stability; within-view accuracy measures agreement with that view's reference judgments.

\begin{table}[!tp]
\centering\footnotesize
\setlength{\tabcolsep}{2.5pt}
\caption{Diagnosis across execution models and evidence views.}
\label{tab:e3_main}
\begin{tabularx}{\linewidth}{@{}>{\raggedright\arraybackslash}Xlr*{4}{rr}@{}}
\toprule
\appendixheader
Benchmark & Executor & $n$ & \multicolumn{2}{c}{Generic LLM} & \multicolumn{2}{c}{Contract LLM} & \multicolumn{2}{c}{Rules} & \multicolumn{2}{c}{\textbf{CAVERT}}\\
\cmidrule(lr){4-5}\cmidrule(lr){6-7}\cmidrule(lr){8-9}\cmidrule(l){10-11}
 & & & Acc. & F1 & Acc. & F1 & Acc. & F1 & Acc. & F1\\
\midrule
\multicolumn{11}{@{}l}{\textit{A. Cross-executor diagnosis: 720 executions}}\\
\addlinespace[2pt]
MARBLE & DeepSeek & 60 & 63.3 & 0.435 & 80.0 & 0.742 & 78.3 & 0.731 & \textbf{86.7} & \textbf{0.828}\\
 & Gemini & 60 & 60.0 & 0.412 & 76.7 & 0.708 & 75.0 & 0.695 & \textbf{83.3} & \textbf{0.792}\\
 & Qwen & 60 & 61.7 & 0.421 & 78.3 & 0.725 & 76.7 & 0.714 & \textbf{85.0} & \textbf{0.810}\\
 & Llama & 60 & 66.7 & 0.468 & 83.3 & 0.785 & 81.7 & 0.768 & \textbf{90.0} & \textbf{0.865}\\
\midrule
CoffeeBench & DeepSeek & 60 & 68.3 & 0.479 & 85.0 & 0.802 & 83.3 & 0.785 & \textbf{90.0} & \textbf{0.865}\\
 & Gemini & 60 & 61.7 & 0.425 & 78.3 & 0.728 & 76.7 & 0.710 & \textbf{85.0} & \textbf{0.812}\\
 & Qwen & 60 & 63.3 & 0.438 & 80.0 & 0.745 & 78.3 & 0.732 & \textbf{86.7} & \textbf{0.830}\\
 & Llama & 60 & 65.0 & 0.452 & 81.7 & 0.762 & 80.0 & 0.748 & \textbf{88.3} & \textbf{0.848}\\
\midrule
MECoBench & DeepSeek & 60 & 66.7 & 0.462 & 81.7 & 0.765 & 80.0 & 0.745 & \textbf{88.3} & \textbf{0.846}\\
 & Gemini & 60 & 65.0 & 0.448 & 80.0 & 0.742 & 78.3 & 0.728 & \textbf{86.7} & \textbf{0.828}\\
 & Qwen & 60 & 65.0 & 0.450 & 81.7 & 0.760 & 80.0 & 0.742 & \textbf{88.3} & \textbf{0.845}\\
 & Llama & 60 & 63.3 & 0.432 & 80.0 & 0.738 & 78.3 & 0.722 & \textbf{86.7} & \textbf{0.825}\\
\midrule
\rowcolor{tableaccent!8}
\multicolumn{2}{l}{\textbf{All benchmarks / executors}} & 720 & 64.2 & 0.435 & 80.6 & 0.748 & 78.9 & 0.738 & \textbf{87.1} & \textbf{0.835}\\
\midrule
\multicolumn{11}{@{}l}{\textit{B. Evidence transformations: 180 traces, five views}}\\
\addlinespace[2pt]
\multicolumn{2}{l}{Original} & 180 & 58.9 & 0.414 & 82.2 & 0.845 & 80.0 & 0.829 & \textbf{87.8} & \textbf{0.895}\\
\multicolumn{2}{l}{Recode A} & 180 & 56.7 & 0.399 & 80.6 & 0.833 & 80.0 & 0.829 & \textbf{87.8} & \textbf{0.895}\\
\multicolumn{2}{l}{Recode B} & 180 & 53.9 & 0.379 & 76.1 & 0.792 & 78.9 & 0.821 & \textbf{86.7} & \textbf{0.886}\\
\multicolumn{2}{l}{Missing L1} & 180 & 41.1 & 0.311 & 68.3 & 0.709 & 65.6 & 0.683 & \textbf{78.3} & \textbf{0.801}\\
\multicolumn{2}{l}{Missing L2} & 180 & 23.9 & 0.217 & 56.1 & 0.568 & 57.2 & 0.578 & \textbf{66.1} & \textbf{0.661}\\
\bottomrule
\end{tabularx}
\tabnote{Acc. is percent; F1 is three-class macro-F1. Upper: 720 executions across three benchmarks and four executors. Lower: a separate 180-trace set under five evidence views; all metrics use the same records.}
\end{table}

\begin{table}[!tp]
\centering\footnotesize
\caption{E3: Joint prediction transitions from per-trace records.}
\label{tab:e3_transitions}
\begin{tabular*}{\linewidth}{@{\extracolsep{\fill}}lrrrrrr@{}}
\toprule
\appendixheader
View & Pred U & To U & From U & N/V flip & Same & Assert U\\
\midrule
\multicolumn{7}{l}{\textit{Generic LLM}}\\
Original & 0 & 0 & 0 & 0 & 180 & 18\\
Recode A & 0 & 0 & 0 & 74 & 106 & 18\\
Recode B & 0 & 0 & 0 & 73 & 107 & 18\\
Missing L1 & 0 & 0 & 0 & 78 & 102 & 38\\
Missing L2 & 0 & 0 & 0 & 70 & 110 & 76\\
\multicolumn{7}{l}{\textit{Contract LLM}}\\
Original & 15 & 0 & 0 & 0 & 180 & 3\\
Recode A & 15 & 3 & 3 & 51 & 123 & 3\\
Recode B & 14 & 2 & 3 & 60 & 115 & 4\\
Missing L1 & 26 & 15 & 4 & 61 & 100 & 12\\
Missing L2 & 43 & 35 & 7 & 57 & 81 & 33\\
\multicolumn{7}{l}{\textit{Strong rules}}\\
Original & 15 & 0 & 0 & 0 & 180 & 3\\
Recode A & 15 & 0 & 0 & 0 & 180 & 3\\
Recode B & 15 & 2 & 2 & 58 & 118 & 3\\
Missing L1 & 25 & 15 & 5 & 62 & 98 & 13\\
Missing L2 & 44 & 35 & 6 & 61 & 78 & 32\\
\multicolumn{7}{l}{\textit{CAVERT}}\\
Original & 16 & 0 & 0 & 0 & 180 & 2\\
Recode A & 16 & 0 & 0 & 0 & 180 & 2\\
Recode B & 16 & 1 & 1 & 32 & 146 & 2\\
Missing L1 & 30 & 18 & 4 & 47 & 111 & 8\\
Missing L2 & 51 & 41 & 6 & 55 & 78 & 25\\
\bottomrule
\end{tabular*}
\tabnote{Transitions compare each view with Original on the same 180 traces. Pred U counts U outputs; To/From U counts N/V-to-U and U-to-N/V changes; N/V flip counts N-to-V or V-to-N changes; Same counts unchanged predictions. Assert U counts definite answers on that view's reference-U cases (18 in lossless views, 38 in L1, 76 in L2).}
\end{table}

\FloatBarrier
\subsection{Prediction Stability and Evidence Sufficiency}
\label{app:l2_errors}

\begin{table}[!tp]
\centering\footnotesize
\caption{Prediction stability under recoding and judgments after evidence removal.}
\label{tab:e3_error_cases}
\begin{tabular*}{\linewidth}{@{\extracolsep{\fill}}lrrl@{}}
\toprule
\appendixheader
Pattern & CAVERT & Rules & CAVERT example\\
\midrule
\multicolumn{4}{@{}l}{\textit{Original to Recode B: paired predictions on 180 traces}}\\
Unchanged prediction & 146/180 & 118/180 & ---\\
Correct to wrong & 18/180 & 32/180 & \texttt{trace\_marble\_016}\\
Wrong to correct & 16/180 & 30/180 & \texttt{trace\_marble\_017}\\
\midrule
\multicolumn{4}{@{}l}{\textit{Within Missing L2: prediction against the remaining-evidence reference}}\\
Reference U, predicted U & 51/76 & 44/76 & ---\\
Reference U, predicted N & 25/76 & 32/76 & \texttt{trace\_coffeebench\_047}\\
Reference V, predicted N & 18/52 & 23/52 & \texttt{trace\_marble\_004}\\
Reference N, predicted V & 18/52 & 22/52 & \texttt{trace\_marble\_054}\\
\bottomrule
\end{tabular*}
\tabnote{Recode B retains Original's reference labels. Missing L2 has 76 U, 52 V, and 52 N references; denominators are the corresponding class sizes. Its three error rows exhaust the observed L2 errors: 61 for CAVERT and 77 for Rules. Recoding transitions and L2 errors concern separate views of the same traces and are not added together.}
\end{table}
\begin{table}[!tp]
\centering\footnotesize
\caption{Paired accuracy differences, CAVERT minus rules.}
\label{tab:e3_paired}
\begin{tabular*}{\linewidth}{@{\extracolsep{\fill}}lrr@{}}
\toprule
\appendixheader
View & Difference (pp) & 95\% interval\\
\midrule
Original & +7.8 & [+5.0, +10.6]\\
Recode A & +7.8 & [+5.0, +10.6]\\
Recode B & +7.8 & [+4.4, +11.7]\\
Missing L1 & +12.8 & [+10.0, +15.6]\\
Missing L2 & +8.9 & [+6.1, +11.7]\\
\bottomrule
\end{tabular*}
\tabnote{10,000 bootstrap draws stratified by benchmark, conditional on the recorded 30-cluster partition of traces. Methods and views share draws.}
\end{table}

Missing L2 tests recognition of insufficient evidence and detection of violations supported by the retained records. Table~\ref{tab:e3_error_cases} reports U-to-N and V-to-N errors alongside false violation predictions (N-to-V).
\begin{appbox}{Missing L2: judgments from the retained evidence}
Rows are reference labels and columns are predictions, both ordered V,N,U:
\[
M_{\mathrm{CAVERT}}=\begin{pmatrix}34&18&0\\18&34&0\\0&25&51\end{pmatrix},\qquad
M_{\mathrm{Rules}}=\begin{pmatrix}29&23&0\\22&30&0\\0&32&44\end{pmatrix}.
\]
\end{appbox}
CAVERT's 25 U-to-N errors comprise five CoffeeBench and twenty MECoBench traces; its 18 V-to-N errors are all from MARBLE. Sixteen of these 18 violations were correctly predicted in the original view. For example, \texttt{trace\_marble\_004} retains reference V but changes from predicted V in L1 to N in L2. CAVERT correctly classifies 119/180 traces and Rules 103/180. These class-specific counts evaluate both evidence sufficiency and duty fulfillment.

\FloatBarrier
\subsection{Diagnosis under a Fixed Spending Limit}
\label{app:budget_diagnosis}

\begin{appbox}{Shared diagnostic inputs and spending limit}
We compare CAVERT with Contract LLM + review on the same 720 E3 executions, public logs, and task contracts. Each benchmark contains four executors and 60 executions per executor. Contract LLM + review extends the single-round Contract LLM with an initial judgment, contract-by-contract review, and revision. Both methods have a per-execution diagnostic spending limit of $B=\$0.0120$, selected on an independent development set and fixed before test evaluation. Stage 1 uses Qwen3.5-Flash; Stage 2 and all Contract LLM calls use DeepSeek-V4-Flash. The limit covers the entire diagnostic process, including retries; CAVERT's final verdict rules require no LLM call. The spending limit uses a fixed schedule per million input/output tokens: $\$0.14/\$0.28$ for DeepSeek-V4-Flash and $\$0.05/\$0.15$ for Qwen3.5-Flash.
\end{appbox}

Every execution remains in the evaluation denominator. Budget exhaustion or an invalid final response is recorded as a diagnostic failure, distinct from the semantic verdict U. A failure receives no accuracy credit and contributes a false negative for its reference class in N/V/U macro-F1. A valid U judgment instead denotes insufficient execution evidence.

CAVERT correctly diagnoses 627/720 executions, compared with 602/720 for Contract LLM + review (Table~\ref{tab:budget_controlled_diagnosis}). Its accuracy advantage is +3.47 pp with a 95\% interval of [+1.39, +5.56]; its macro-F1 advantage is +0.049 [+0.021, +0.076]. Both point estimates favor CAVERT in every benchmark (Table~\ref{tab:budget_diagnosis_intervals}). The comparison extends the practical benefit of the criteria to a baseline with iterative contract review under the same spending limit. CAVERT averages 1.81 model calls per execution, versus 2.65 for the review baseline. Mean API cost is $\$0.0017$ for CAVERT and $\$0.0031$ for Contract LLM + review, with 95th-percentile costs of $\$0.0026$ and $\$0.0042$, respectively (Table~\ref{tab:diagnostic_cost_audit}). The diagnostic advantage thus persists with iterative review available to the baseline, while CAVERT uses fewer calls and lower API spending under the fixed price schedule.

For uncertainty estimates, 10,000 paired bootstrap replicates resample original task clusters within each benchmark. Each sampled cluster retains its executions, executor configurations, and both methods' predictions. Accuracy and three-class macro-F1 are recomputed from pooled predictions in each replicate. The 95\% intervals are the 2.5th and 97.5th percentiles of paired differences. Pooled macro-F1 is computed across executions; it is not the mean of benchmark macro-F1 scores.

\begin{table}[!tp]
\centering\footnotesize
\caption{Diagnosis under a common per-execution spending limit of $\$0.0120$.}
\label{tab:budget_controlled_diagnosis}
\begin{tabular*}{\linewidth}{@{\extracolsep{\fill}}llrrrr@{}}
\toprule
\appendixheader
Benchmark & Method & Correct & Acc. (\%) & Macro-F1 & Calls\\
\midrule
MARBLE & Contract LLM + review & 198/240 & 82.50 & 0.778 & 2.64\\
MARBLE & \textbf{CAVERT} & \textbf{207/240} & \textbf{86.25} & \textbf{0.824} & \textbf{1.80}\\
\addlinespace[2pt]
CoffeeBench & Contract LLM + review & 203/240 & 84.58 & 0.801 & 2.66\\
CoffeeBench & \textbf{CAVERT} & \textbf{210/240} & \textbf{87.50} & \textbf{0.839} & \textbf{1.82}\\
\addlinespace[2pt]
MECoBench & Contract LLM + review & 201/240 & 83.75 & 0.779 & 2.65\\
MECoBench & \textbf{CAVERT} & \textbf{210/240} & \textbf{87.50} & \textbf{0.841} & \textbf{1.81}\\
\midrule
Pooled & Contract LLM + review & 602/720 & 83.61 & 0.786 & 2.65\\
Pooled & \textbf{CAVERT} & \textbf{627/720} & \textbf{87.08} & \textbf{0.835} & \textbf{1.81}\\
\bottomrule
\end{tabular*}
\tabnote{Each benchmark aggregates four executors $\times$ 60 executions. Calls are per-execution means over the complete diagnostic process, including format-repair attempts and retries. Both methods may finish below the spending limit. Single-round Contract LLM results are reported separately in Table~\ref{tab:diagnosis_summary}.}
\end{table}

\begin{table}[!tp]
\centering\footnotesize
\caption{Paired performance differences under the common spending limit.}
\label{tab:budget_diagnosis_intervals}
\begin{tabular*}{\linewidth}{@{\extracolsep{\fill}}lrrrr@{}}
\toprule
\appendixheader
Scope & $\Delta$ Acc. (pp) & 95\% interval & $\Delta$ Macro-F1 & 95\% interval\\
\midrule
MARBLE & +3.75 & [+0.42, +7.08] & +0.046 & [+0.012, +0.080]\\
CoffeeBench & +2.92 & [+0.00, +5.83] & +0.038 & [+0.006, +0.070]\\
MECoBench & +3.75 & [+0.83, +6.67] & +0.062 & [+0.025, +0.099]\\
Pooled & +3.47 & [+1.39, +5.56] & +0.049 & [+0.021, +0.076]\\
\bottomrule
\end{tabular*}
\tabnote{Differences are CAVERT minus Contract LLM + review. Intervals use 10,000 paired task-cluster bootstrap replicates stratified by benchmark.}
\end{table}

\begin{table}[!tp]
\centering\footnotesize
\caption{API usage and cost across all 720 E3 executions.}
\label{tab:diagnostic_cost_audit}
\begin{tabularx}{\linewidth}{@{}lYrrr@{}}
\toprule
\appendixheader
Method & Diagnostic path & Calls & Mean cost & P95 cost\\
\midrule
Generic LLM & Direct judgment & 1.00 & $\$0.0010$ & $\$0.0014$\\
Rules & Deterministic checks & 0.00 & $\$0.0000$ & $\$0.0000$\\
Contract LLM & Single-round contract judgment & 1.00 & $\$0.0014$ & $\$0.0021$\\
Contract LLM + review & Initial judgment, review, revision & 2.65 & $\$0.0031$ & $\$0.0042$\\
\textbf{CAVERT} & Scope filtering, causal-state inference, deterministic verdict & \textbf{1.81} & \textbf{\$0.0017} & \textbf{\$0.0026}\\
\bottomrule
\end{tabularx}
\tabnote{Calls and costs cover each execution's complete diagnostic process, including format-repair attempts and billed failed calls or retries. Means use all 720 executions; P95 is the 95th percentile of per-execution total cost. USD costs use recorded token consumption and the fixed price schedule in the text. Per-call output ceilings are not consumption measurements. Rules incur no model API cost.}
\end{table}

\FloatBarrier
\section{Execution Protocol and Stratification}
\label{app:e4e6}\label{sec:rq3_auxiliary}

\FloatBarrier
\subsection{Matched Blocks and Executor Configurations}
\label{app:recovery_config}

E4 and E5 share 720 fault blocks but score complete executions and truncated probes separately. E5 reports 6,480 base nine-arm probes, 720 fixed-step Rules+G sensitivity probes, and 576 probes from four length configurations on 144 edge-case blocks. Step-0 logs are shared, so E4 and E5 reported run counts are not simply added.

E4 assigns 90 blocks per benchmark to each executor configuration. Config-A uses DeepSeek-V4-Flash, temperature 0.2, top-$p=0.95$, a 2,048-token output ceiling, a 30-action budget, and a 180-second timeout. Config-B uses Gemini-2.5-Flash, temperature 0, top-$p=1$, a 4,096-token output ceiling, a 50-action budget, and a 300-second timeout. Their controllers are the standard and extended variants of \texttt{cavert\_controller\_v4.2}. The original allocation has six arms: Native, Retry, Direct+G, Rules+G, CAVERT-Prompt, and CAVERT-Gate. Supplementary MemTX evaluation adds one arm on the same 720 blocks, retaining each block's executor configuration and resource ceilings. Table~\ref{tab:e4_execution} and Figure~\ref{fig:recovery_outcomes} display the six intervention arms. The Native score of 27.8\% in Table~\ref{tab:nine_arms} belongs to E5's fixed-length probes, not E4's complete-execution endpoint.

\begin{table}[!tp]
\centering\footnotesize
\setlength{\tabcolsep}{2pt}
\caption{Matched complete-execution recovery with benchmark-specific endpoints.}
\label{tab:e4_execution}
\begin{tabularx}{\linewidth}{@{}lYrrrrrrr@{}}
\toprule
\appendixheader
 & & & \multicolumn{4}{c}{Baselines} & \multicolumn{2}{c}{\textbf{CAVERT}}\\
\cmidrule(lr){4-7}\cmidrule(l){8-9}
Benchmark & Endpoint & $n$/arm & Retry & Direct+G & Rules+G & MemTX & Prompt & Gate\\
\midrule
\multicolumn{9}{@{}l}{\textit{Matched complete executions: 180 blocks per benchmark}}\\
AppWorld & Success \% & 180 & 31.1 & 38.9 & 49.4 & 55.0 & 43.3 & \textbf{67.8}\\
LongCLI & F2P+P2P \% & 180 & 12.2 & 18.9 & 28.3 & 32.8 & 22.2 & \textbf{41.7}\\
MARBLE & Score \% & 180 & 46.7 & 53.9 & 64.4 & 72.2 & 57.8 & \textbf{78.3}\\
CoffeeBench & Payout \% & 180 & 43.9 & 51.1 & 62.2 & 68.9 & 55.0 & \textbf{76.1}\\
\midrule
\rowcolor{tableaccent!8}
\multicolumn{2}{@{}l}{Equal-weight normalized mean} & 720 & 33.5 & 40.7 & 51.1 & 57.2 & 44.6 & \textbf{66.0}\\
\bottomrule
\end{tabularx}
\tabnote{G: shared gate. All rows use two executor configurations, with 90 blocks each and 180 per arm (Appendix~\ref{app:e4e6}). MemTX adds one arm on the same blocks. Payout is clipped normalization, not dollar income. The final row averages four normalized endpoint scores, not task-success probabilities across benchmarks. Paired differences appear in Table~\ref{tab:memtx_recovery_differences}.}
\end{table}

Matched executor limits refer to the action budget, model-output ceiling, and timeout assigned to a block. Diagnostic calls have their own per-call output ceilings (Table~\ref{tab:delivery_config}). Task-level token consumption, latency, and billed cost describe realized overhead and are reported separately in Table~\ref{tab:recovery_overhead}.

MemTX~\citep{li2026memtx} is a complete memory-system comparator, with belief commit, action gating, and cascading repair. Table~\ref{tab:memtx_recovery_differences} compares its recovery scores with Rules+G and CAVERT-Gate. For each of 10,000 percentile-bootstrap replicates, we resample matched blocks within each benchmark, preserving task, executor configuration, and seed across arms. The aggregate is the equal-weight mean of the four benchmark differences. CAVERT-Gate improves on MemTX in every benchmark, with an aggregate difference of +8.75 pp [6.39,11.11]. MemTX improves on Rules+G by +6.11 pp [3.89,8.33] overall; its LongCLI interval includes zero. These are comparisons of complete intervention configurations.

\begin{table}[!tp]
\centering\small
\caption{Paired recovery score differences on the same 720 fault blocks.}
\label{tab:memtx_recovery_differences}
\begin{tabularx}{\linewidth}{@{}Yrr@{}}
\toprule
\appendixheader
Benchmark & MemTX $-$ Rules+G & CAVERT-Gate $-$ MemTX\\
\midrule
AppWorld & $+5.56\ [1.11,10.00]$ & $+12.78\ [7.78,17.78]$\\
LongCLI & $+4.44\ [-0.56,9.44]$ & $+8.89\ [3.89,13.89]$\\
MARBLE & $+7.78\ [3.33,12.22]$ & $+6.11\ [1.67,10.56]$\\
CoffeeBench & $+6.67\ [2.22,11.11]$ & $+7.22\ [2.78,11.67]$\\
\midrule
\rowcolor{tableaccent!8}
Equal-weight mean difference & $+6.11\ [3.89,8.33]$ & $+8.75\ [6.39,11.11]$\\
\bottomrule
\end{tabularx}
\tabnote{Entries are differences in percentage points with 95\% intervals, computed before rounding scores. The 10,000 percentile-bootstrap replicates resample matched blocks within each benchmark, keeping task, executor configuration, and seed paired across arms. Each aggregate replicate averages the four benchmark differences with equal weight. Intervals are pointwise.}
\end{table}

\FloatBarrier
\subsection{Recovery Controls and Outcome Definitions}
\label{app:recovery_endpoints}

\begin{table}[!tp]
\centering\small
\caption{E5: Nine-arm suffix-probe results.}
\label{tab:nine_arms}
\begin{tabular*}{\linewidth}{@{\extracolsep{\fill}}lr@{}}
\toprule
\appendixheader
Arm & Macro-average score (\%)\\
\midrule
native & 27.8\\
retry & 33.5\\
generic prompt & 36.9\\
rules prompt & 42.5\\
direct prompt & 43.1\\
cavert prompt & 48.3\\
rules gate & 51.1\\
direct gate & 52.2\\
cavert gate & 66.0\\
\bottomrule
\end{tabular*}
\end{table}

\begin{table}[!tp]
\centering
\begin{minipage}{0.84\linewidth}
\centering\footnotesize
\caption{Diagnosis--intervention interaction under fixed-length suffix evaluation.}
\label{tab:e5_interaction}
\begin{tabularx}{\linewidth}{@{}l*{4}{>{\raggedleft\arraybackslash}X}@{}}
\toprule
\multicolumn{5}{@{}l}{\textbf{720 matched blocks; four-benchmark mean}}\\
Judge & Prompt (\%) & Gate (\%) & Gain (pp) & 95\% paired CI\\
\midrule
Strong rules & 42.5 & 51.1 & $+8.6$ & $[5.2,12.0]$\\
Direct LLM & 43.1 & 52.2 & $+9.1$ & $[5.6,12.6]$\\
\rowcolor{tableaccent!9}
CAVERT & 48.3 & 66.0 & $+17.7$ & $[13.8,21.6]$\\
\bottomrule
\end{tabularx}
\par\smallskip
\tabnote{AppWorld, LongCLI, MARBLE, and CoffeeBench contribute 180 blocks each; judges share the gate. Gain is gate minus prompt. CAVERT versus rules interaction: +9.1 pp, 95\% CI [4.8,13.4], $p<0.001$. Scores average benchmark-specific normalized endpoints, not pooled success probabilities.}
\end{minipage}
\end{table}

Native, retry, and generic-prompt controls accompany rules, direct LLM, and CAVERT crossed with prompt/gate. Arms share initial prefixes, action interfaces, and executor limits within a block. The gate intercepts writes before commitment.

\begin{appbox}{Recovery endpoints and aggregate score}
AppWorld uses binary official success; LongCLI uses joint F2P/P2P pass; MARBLE uses a continuous task score. CoffeeBench uses the normalization
\[
s_{\rm CB}=\min(1,\max(0,\mathrm{Payout}/500)),\qquad
\bar s=\tfrac14\sum_{b=1}^{4}\bar s_b.
\]
Payout is cumulative total earnings within a 30-environment-action scoring window, fixed across arms and executor configurations. The preregistered custom clipping threshold, $\mathrm{Payout}_{\rm trunc}=\$500$, uses the reported 95th-percentile threshold of historical market scenarios. All arms share this normalization and zero floor. This windowed earnings score is reported separately from cumulative net income; the executor's action ceiling controls the run budget rather than the scoring window.
\end{appbox}

\FloatBarrier
\subsection{Normal-Task Impact and Budget Sensitivity}
\label{app:normal_uncertainty}

\begin{table}[!tp]
\centering\footnotesize
\caption{Budget sensitivity and interference on normal tasks.}
\label{tab:e6_conditions}
\begin{tabularx}{\linewidth}{@{}Xrrr@{}}
\toprule
\multicolumn{4}{@{}l}{\textit{A. Budget sensitivity: 240 fault blocks}}\\
\appendixheader
\shortstack{Actions /\\tokens} & \shortstack{Generic\\Prompt} & Rules+G & CAVERT+G\\
\midrule
15 / 50k & 28.3 & 41.7 & 54.2\\
30 / 100k & 36.9 & 51.1 & 66.0\\
50 / 200k & 40.8 & 55.4 & 71.3\\
\bottomrule
\end{tabularx}
\par\medskip
\begin{tabularx}{\linewidth}{@{}Yrrrl@{}}
\toprule
\multicolumn{5}{@{}l}{\textit{B. Normal tasks: 480 uninjected blocks}}\\
\appendixheader
Method & Score & $\Delta$ (pp) & 95\% CI & Non-inferiority\\
\midrule
Native & 88.5 & $+0.0$ & Reference & --\\
Rules+G & 86.1 & $-2.4$ & $[-4.1,-0.7]$ & Not established\\
MemTX (mature) & 86.7 & $-1.8$ & $[-3.4,-0.2]$ & Not established\\
MemTX (unmatured) & 84.6 & $-3.9$ & $[-5.6,-2.1]$ & Not established\\
\rowcolor{tableaccent!9}
CAVERT+G & 87.8 & $-0.7$ & $[-2.1,+0.7]$ & Supported\\
\bottomrule
\end{tabularx}
\par\smallskip
\tabnote{Scores are percentages. A reuses a matched-fault subset; action/token values are ceilings. B compares interventions with Native on the same 480 blocks. Non-inferiority requires the 95\% interval's lower bound to exceed $-2.5$ pp (paired-block bootstrap; Appendix~\ref{app:normal_uncertainty}).}
\end{table}

Table~\ref{tab:e6_strata} reports normal-task scores for four 120-block cohorts. CAVERT meets the $-2.5$ pp non-inferiority criterion for the 480-block aggregate; the MARBLE stratum's lower bound equals the margin. On the same blocks, mature and unmatured MemTX score 86.7\% and 84.6\%, respectively. Each is compared with Native: $-1.8$ pp [$-3.4,-0.2$] and $-3.9$ pp [$-5.6,-2.1$]. Both intervals cross the non-inferiority margin.

The 51-block $\tau^2$-bench~\citep{barres2025tau2bench} action-audit cohort is separate. Of its 750 permitted actions, Rules+G blocks 43, MemTX 29, and CAVERT+G 12 (Table~\ref{tab:recovery_overhead}). These action-level rates are distinct from four-benchmark task scores.

For normal-task uncertainty, we resample the 480 matched blocks with replacement 10,000 times, retaining each block's arm scores together. The 2.5th and 97.5th percentiles of the mean CAVERT-minus-Native difference give $[-2.1,+0.7]$ pp around $-0.7$ pp. Non-inferiority requires the lower bound to exceed the predeclared $-2.5$ pp margin.

\begin{table}[!tp]
\centering\footnotesize
\setlength{\tabcolsep}{4pt}
\caption{Normal-task scores and paired differences from Native, by benchmark.}
\label{tab:e6_strata}
\begin{tabularx}{\linewidth}{@{}Yrrrrr@{}}
\toprule
\multicolumn{6}{@{}l}{\textit{A. Scores (\%); MemTX uses mature initialization}}\\
\appendixheader
Cohort & Blocks & Native & Rules+G & MemTX & CAVERT+G\\
\midrule
AppWorld & 120 & 90.0 & 87.5 & 88.3 & 89.2\\
LongCLI & 120 & 85.0 & 82.5 & 83.3 & 84.2\\
MARBLE & 120 & 89.2 & 86.7 & 87.5 & 88.3\\
CoffeeBench & 120 & 89.8 & 87.8 & 87.8 & 89.4\\
\midrule
\rowcolor{tableaccent!8}
Equal-weight mean & 480 & 88.5 & 86.1 & 86.7 & 87.8\\
\midrule
Historical $\tau^2$ & 51 & 88.2 & 85.8 & -- & 87.5\\
\bottomrule
\end{tabularx}
\par\medskip
\begin{tabularx}{\linewidth}{@{}Yrr@{}}
\toprule
\multicolumn{3}{@{}l}{\textit{B. Differences from Native (pp), with 95\% intervals}}\\
\appendixheader
Cohort & MemTX $-$ Native & CAVERT+G $-$ Native\\
\midrule
AppWorld & $-1.7\ [-3.3,+0.0]$ & $-0.8\ [-2.3,+0.7]$\\
LongCLI & $-1.7\ [-3.4,+0.1]$ & $-0.8\ [-2.4,+0.8]$\\
MARBLE & $-1.7\ [-3.4,+0.1]$ & $-0.9\ [-2.5,+0.7]$\\
CoffeeBench & $-2.0\ [-3.5,-0.5]$ & $-0.4\ [-1.8,+1.0]$\\
\midrule
\rowcolor{tableaccent!8}
Equal-weight mean & $-1.8\ [-3.4,-0.2]$ & $-0.7\ [-2.1,+0.7]$\\
\midrule
Historical $\tau^2$ & -- & $-0.7\ [-2.1,+0.7]$\\
\bottomrule
\end{tabularx}
\tabnote{Four 120-block cohorts form the 480-block aggregate. The historical 51 are separate; no MemTX task score is reported for that cohort. Its permitted-action comparison appears in Table~\ref{tab:recovery_overhead}. Normal-task scores and action false-block rates have different denominators.}
\end{table}

\begin{figure}[!tp]
\centering
\includegraphics[width=\linewidth]{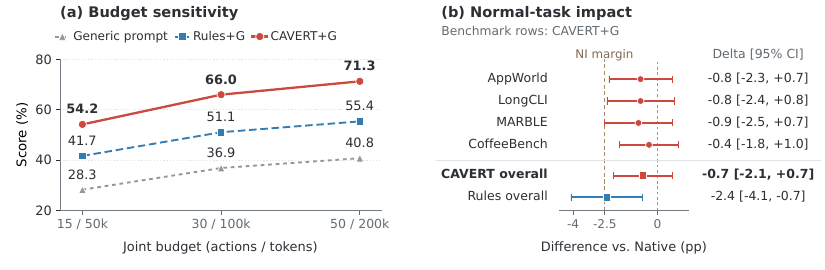}
\caption{Budget sensitivity (a, 240 fault blocks) and paired normal-task differences from Native (b, 480 blocks). Benchmark rows show CAVERT+G with 95\% CIs. Dashed lines mark zero and the $-2.5$ pp margin. Protocol: Tables~\ref{tab:e6_conditions} and~\ref{tab:e6_strata}.}
\label{fig:budget_normal_detail}
\end{figure}

Budget tiers jointly change action ceilings (15, 30, 50) and token ceilings (50k, 100k, 200k). They therefore measure combined budget sensitivity, not an isolated token effect.

\FloatBarrier
\subsection{Supplementary Cohorts, Outcomes, and Costs}
\label{app:prior_rq3}
\begin{table}[!tp]
\centering\footnotesize
\setlength{\tabcolsep}{2pt}
\caption{Recovery across execution models and supplementary cohorts.}
\label{tab:rq3_model_details}
\begin{tabularx}{\linewidth}{@{}ll>{\raggedright\arraybackslash}Xrrrrrr@{}}
\toprule
\appendixheader
 & & & & \multicolumn{3}{c}{Baselines} & \multicolumn{2}{c}{\textbf{CAVERT}}\\
\cmidrule(lr){5-7}\cmidrule(l){8-9}
Benchmark & Executor & Endpoint & $n$/arm & Retry & Direct+G & Rules+G & Prompt & Gate\\
\midrule
\multicolumn{9}{@{}l}{\textit{Cross-backbone recovery}}\\
AppWorld & DeepSeek & Recovery \% & 120 & 25.8 & 34.2 & 51.7 & 36.7 & \textbf{68.3}\\
 & Gemini &  & 120 & 26.7 & 35.0 & 52.5 & 37.5 & \textbf{69.2}\\
 & Qwen &  & 120 & 20.8 & 29.2 & 45.8 & 30.8 & \textbf{61.7}\\
 & Llama &  & 120 & 22.5 & 30.8 & 47.5 & 32.5 & \textbf{64.2}\\
\midrule
LongCLI & DeepSeek & F2P+P2P \% & 60 & 6.7 & 13.3 & 28.3 & 15.0 & \textbf{43.3}\\
 & Gemini &  & 60 & 6.7 & 13.3 & 28.3 & 15.0 & \textbf{43.3}\\
 & Qwen &  & 60 & 5.0 & 11.7 & 25.0 & 13.3 & \textbf{38.3}\\
 & Llama &  & 60 & 6.7 & 13.3 & 26.7 & 15.0 & \textbf{40.0}\\
\midrule
\multicolumn{9}{@{}l}{\textit{Original and extended cohorts: native endpoints}}\\
\rowcolor{black!4}
AppWorld & 4-model & Success \% & 480 & 18.33 & 25.00 & 40.83 & 27.50 & \textbf{54.17}\\
\rowcolor{black!4}
LongCLI & 4-model & F2P+P2P \% & 240 & 6.25 & 12.92 & 27.08 & 14.58 & \textbf{41.25}\\
MARBLE & Llama & Score \% & 64 & 54.2 & 67.8 & 76.5 & 73.5 & \textbf{88.5}\\
CoffeeBench & DeepSeek & Income (USD) & 18 & 1,240 & 2,420 & 3,580 & 3,150 & \textbf{4,680}\\
EntCollabBench & DS/Gemini & Recovery \% & 24 & 33.3 & -- & -- & 54.2 & \textbf{79.2}\\
PAST-Bench & Qwen/Llama & Score [0,1] & 90 & 0.583* & -- & -- & -- & \textbf{0.867*}\\
EntCollab$^{\ddagger}$ & DeepSeek & Workflow \% & 180 & 37.8 & 48.3 & 58.9 & 45.6 & \textbf{72.2}\\
\bottomrule
\end{tabularx}
\tabnote{G: shared gate. AppWorld recovery flags and official success are distinct endpoints. * PAST uses persistence/controller interfaces (Table~\ref{tab:team_additional_recovery}). EntCollab (24 groups): DeepSeek-V4-Flash/Gemini-2.5-Flash; $\ddagger$: independent extension. Dollar income is unclipped. Cohort settings: Appendix~\ref{app:prior_rq3}.}
\end{table}

The supplementary AppWorld cohort contains 40 task families, four models, and three seeds (480 blocks per arm). LongCLI contains 20 tasks, four models, and three seeds (240 blocks per arm). Both allow 15 interaction steps: 11 prefix steps and four recovery steps. This setting is separate from the budget sweep. AppWorld scores recovery sequencing, state postconditions, and official success separately: CAVERT+G obtains 316/480 recovery flags and 260/480 official successes. Paired resampling retains AppWorld families and repository-aware LongCLI groups with all models, seeds, and arms together. MARBLE uses task groups and CoffeeBench uses worlds. The CoffeeBench dollar-income panel reports unclipped income, separately from the normalized score panel.

Our EntCollabBench~\citep{yu2026entcollab} cohort has 24 recovery groups, six per fault family, within the original 36 semantic groups. Success requires official task passage and a final database state matching the reference. The 24 groups across three arms give 72 runs, spanning DeepSeek-V4-Flash and Gemini-2.5-Flash. Retry, prompt, and gate succeed on 8/24, 13/24, and 19/24 groups, respectively. The gate-minus-prompt difference is +25.0 pp, with 11 gate-only and five prompt-only successes. A two-sided exact McNemar test gives $p=0.2101$; this small-cohort difference is not statistically significant at 0.05.

Our PAST-Bench~\citep{xue2026pastbench} evaluation measures the fraction of subgoals passed over 90 episodes per arm (30 near, 60 far), five families, and three seeds. No persistence, generic persistence, structured records, and contract control score 0.425, 0.583, 0.742, and 0.867. State access and control authority vary jointly. We aggregate episodes and seeds within families, then resample the five families 20,000 times. Models are fixed within families.

The action panel reuses the 750 permitted actions rather than adding independent task blocks. Token increases use a common Native mean of 12,000 tokens per task. MemTX uses 14,220 tokens and 22.3 seconds per task; CAVERT+G uses 14,950 tokens and 23.5 seconds. Their AppWorld/LongCLI costs are \$0.044 and \$0.042 per task under DeepSeek-V4-Flash billing, including prompt-cache discounts. Token totals alone therefore do not determine billed cost.

\begin{table}[!tp]
\caption{Five-arm supplementary recovery: native task outcomes and suffix counts.}
\label{tab:rq3_cross_benchmark}
\label{tab:suffix_endpoint_counts}
\centering
\footnotesize
\begin{tabularx}{\linewidth}{@{}Yrrrr@{}}
\toprule
\appendixheader
\multicolumn{5}{l}{\textbf{A. Native multi-agent complete tasks}}\\
& \multicolumn{2}{c}{MARBLE: 64 tasks} & \multicolumn{2}{c}{CoffeeBench: 18 worlds}\\
\cmidrule(lr){2-3}\cmidrule(l){4-5}
Recovery method & Task score $\uparrow$ & Violations (\%) $\downarrow$ & Income (USD) $\uparrow$ & Violations (\%) $\downarrow$\\
\midrule
Standard retry & 0.542 & 31.6 & 1,240 & 42.3\\
Fresh-read recovery & 0.615 & 25.2 & 1,850 & 33.5\\
Direct diagnosis + gate & 0.678 & 18.4 & 2,420 & 24.8\\
CAVERT prompt-only & 0.735 & 11.2 & 3,150 & 15.6\\
\midrule
\rowcolor{tableaccent!8}
\textbf{CAVERT + gate} & \textbf{0.885} & \textbf{1.8} & \textbf{4,680} & \textbf{2.3}\\
\bottomrule
\end{tabularx}
\smallskip
\begin{tabularx}{\linewidth}{@{}Yrrrrr@{}}
\toprule
\appendixheader
\multicolumn{6}{l}{\textbf{B. Stateful recovery suffixes: counts}}\\
& \multicolumn{3}{c}{AppWorld: 480 blocks/arm} & \multicolumn{2}{c}{LongCLI: 240 blocks/arm}\\
\cmidrule(lr){2-4}\cmidrule(l){5-6}
Recovery method & Recovery flag & Postcondition & Success & F2P+P2P & Regressions\\
\midrule
Standard retry & 115 & 104 & 88 & 15 & 41\\
Fresh-read recovery & 137 & 124 & 108 & 23 & 34\\
Direct diagnosis + gate & 155 & 136 & 120 & 31 & 23\\
CAVERT prompt-only & 165 & 152 & 132 & 35 & 18\\
\midrule
\rowcolor{tableaccent!8}
\textbf{CAVERT + gate} & \textbf{316} & \textbf{292} & \textbf{260} & \textbf{99} & \textbf{4}\\
\bottomrule
\end{tabularx}
\tabnote{Direct uses the same gate; prompt-only uses CAVERT diagnosis without blocking. A: violations per assessed obligation. B: 40 AppWorld families / 20 LongCLI tasks $\times$ 4 models $\times$ 3 seeds. AppWorld scores recovery sequencing, state postconditions, and official success separately. F2P+P2P requires repair without test regressions; fewer regressions is better.}
\end{table}

\begin{table}[!tp]
\caption{Enterprise recovery and cross-session scores under their original intervention interfaces.}
\label{tab:team_additional_recovery}
\centering
\footnotesize
\begin{tabularx}{\linewidth}{@{}L{0.20\linewidth}rrrY@{}}
\toprule
\appendixheader
Panel / endpoint & Reference & Record & Controller & $\Delta_{\rm C-R}$ [95\% CI]\\
\midrule
EntCollabBench: recovery & 0.333 & 0.542 & 0.792 & +0.458 [0.250, 0.667]\\
PAST-Bench: cross-session score & 0.583 & 0.742 & 0.867 & +0.284 [0.183, 0.385]\\
\bottomrule
\end{tabularx}
\tabnote{$\Delta_{\rm C-R}$ is Controller minus Reference before rounding, in rate/score units. EntCollabBench columns: Retry, CAVERT prompt, CAVERT Gate (8/24, 13/24, 19/24). PAST-Bench columns: Generic Persistence, Root/Trigger Record, Contract Controller; 90 episodes per arm and 20,000 paired bootstrap draws over five families.}
\end{table}

\begin{table}[!tp]
\caption{Action false blocks, execution overhead, and monetary costs, with separate cohorts and units.}
\label{tab:recovery_overhead}
\centering
\footnotesize
\begin{tabularx}{\linewidth}{@{}Yrrl@{}}
\toprule
\multicolumn{4}{@{}l}{\textit{A. Historical $\tau^2$-bench: 750 permitted actions}}\\
\appendixheader
Method & False blocks / permitted actions & Rate & 95\% CI\\
\midrule
Native & 0/750 & 0.0\% & --\\
Rules+G & 43/750 & 5.7\% & [4.2\%, 7.5\%]\\
MemTX & 29/750 & 3.9\% & [2.6\%, 5.4\%]\\
\rowcolor{tableaccent!8}
CAVERT+G & 12/750 & 1.6\% & [0.8\%, 2.5\%]\\
\bottomrule
\end{tabularx}
\par\medskip
\begin{tabularx}{\linewidth}{@{}Yrrrr@{}}
\toprule
\multicolumn{5}{@{}l}{\textit{B. Long-horizon execution: per-task means}}\\
\appendixheader
Method & Tokens & Token increase & Latency (s) & Added latency (s)\\
\midrule
Native & 12,000 & Reference & 18.5 & Reference\\
Rules+G & 13,200 & +10.0\% & 19.8 & +1.3\\
MemTX & 14,220 & +18.5\% & 22.3 & +3.8\\
\rowcolor{tableaccent!8}
CAVERT+G & 14,950 & +24.6\% & 23.5 & +5.0\\
\bottomrule
\end{tabularx}
\par\medskip
\begin{tabularx}{\linewidth}{@{}Yrrrr@{}}
\toprule
\multicolumn{5}{@{}l}{\textit{C. AppWorld + LongCLI: DeepSeek-V4-Flash billing}}\\
\appendixheader
Unit & Native & Rules+G & MemTX & CAVERT+G\\
\midrule
USD / task & 0.030 & 0.035 & 0.044 & 0.042\\
\bottomrule
\end{tabularx}
\tabnote{A: separate from the 720 fault and 480 normal blocks; -- denotes no interval. B: per-task means on a separate long-horizon cohort, relative to Native. C: AppWorld/LongCLI billed costs including prompt-cache discounts.}
\end{table}

\FloatBarrier
\subsection{Limitations}
\label{app:limitations}

Our claims are limited to the evaluated contracts, evidence factors, and log transformations; natural-log frequencies characterize only the sampled sources. CAVERT recovers semantic relationships from existing logs rather than observing them natively, so diagnostic quality remains bounded by extraction quality. The recovery gate acts before commitment; repairing committed external effects requires compensation mechanisms beyond our current scope.

\end{document}